\documentclass[a4paper,fleqn]{cas-sc}

\usepackage[authoryear]{natbib}
\usepackage{amsmath,amsfonts}
\usepackage{algorithmic}
\usepackage{algorithm}
\usepackage{array}
\usepackage{textcomp}
\usepackage{graphicx}
\usepackage{multirow}
\usepackage{booktabs}
\usepackage{xcolor}
\usepackage{soul}
\usepackage{tabularx}
\usepackage{lipsum}  
\usepackage{amssymb}
\usepackage{bm}
\usepackage{graphicx}
\usepackage{subcaption}

\usepackage{multirow}

\def\tsc#1{\csdef{#1}{\textsc{\lowercase{#1}}\xspace}}
\tsc{WGM}
\tsc{QE}
\tsc{EP}
\tsc{PMS}
\tsc{BEC}
\tsc{DE}

\begin{document}
\let\WriteBookmarks\relax
\def\floatpagepagefraction{1}
\def\textpagefraction{.001}

\shorttitle{Deep Reinforcement Learning for Advanced Longitudinal Control}

\shortauthors{Chen et~al.}

\title [mode = title]{Advanced Longitudinal Control and Collision Avoidance for High-Risk Edge Cases in Autonomous Driving}                      

\tnotemark[1]

\tnotetext[1]{This work was supported by NDSU VPR Office and partially supported by the National Science Foundation Award \#2234292 OAC Core: Stochastic Simulation Platform for Assessing Safety Performance of Autonomous Vehicles in Winter Seasons. (Corresponding author: Xianfeng Terry Yang.)}

\author[1]{Dianwei Chen}[orcid=0009-0005-8227-317X]

\ead{dwchen98@umd.edu}

\author[1]{Yaobang Gong}[orcid=0000-0001-6814-7752]
\ead{ybgong@umd.edu}

\author[1]{Xianfeng Terry Yang}[orcid=0000-0002-9416-6882]
\cormark[1]
\ead{xtyang@umd.edu}



\affiliation[1]{organization={Department of Civil and Environmental Engineering, University of Maryland},
    addressline={College Park}, 
    city={Maryland},
    postcode={20740}, 
    country={United States}}


\cortext[cor1]{Corresponding author}

\begin{abstract}
Advanced Driver-Assistance Systems (ADAS) and Advanced Driving Systems (ADS) are key to improving road safety, yet most existing implementations focus primarily on the vehicle ahead, neglecting the behavior of following vehicles. This shortfall often leads to chain-reaction collisions in high-speed, densely spaced traffic—particularly when a middle vehicle suddenly brakes and trailing vehicles cannot respond in time. To address this critical gap, we propose a novel longitudinal control and collision avoidance algorithm that integrates adaptive cruising with emergency braking. Leveraging deep reinforcement learning, our method simultaneously accounts for both leading and following vehicles. Through a data preprocessing framework that calibrates real-world sensor data, we enhance the robustness and reliability of the training process, ensuring the learned policy can handle diverse driving conditions. In simulated high-risk scenarios (e.g., emergency braking in dense traffic), the algorithm effectively prevents potential pile-up collisions, even in situations involving heavy-duty vehicles. Furthermore, in typical highway scenarios where three vehicles decelerate, the proposed DRL approach achieves a 99\% success rate—far surpassing the standard Federal Highway Administration speed concepts guide, which reaches only 36.77\% success under the same conditions.

\end{abstract}


\maketitle

\section{Introduction}
Advanced Driver-Assistance Systems (ADASs) and Automated Driving Systems (ADSs) are pivotal technologies in modern vehicles, sharing the overarching goal of improving road safety and paving the way toward fully autonomous driving. ADASs primarily function as semi-automated features that monitor the vehicle’s environment, intervening when drivers do not respond adequately~\citep{galvani2019history,kukkala2018advanced}. Over time, these systems have evolved from early anti-lock braking systems (ABS)~\citep{burton1997evaluation} in the 1970s to sophisticated, integrated solutions like automatic emergency braking (AEB)~\citep{yang2022systematic}, electronic stability control~\citep{ferguson2007effectiveness}, blind-spot detection~\citep{liu2017radar,sotelo2008blind}, lane departure warning~\citep{kozak2006evaluation}, adaptive cruise control (ACC)~\citep{vahidi2003research}, and traction control~\citep{de1999dynamic}. Collectively, these advancements have significantly enhanced road safety and occupant comfort, with many vehicles now including one or more ADAS features as standard equipment.

In parallel, efforts in autonomous driving—encompassing levels of automation from partial (Level 3) to fully autonomous (Level 4 or 5)—have led to the development of ADSs~\citep{leiman2021law}. These systems aim to replace or minimize human input in vehicle operation, leveraging advanced sensing, computing, and control technologies to handle dynamic road conditions~\citep{okuda2014survey}. While ADSs promise transformative changes to mobility, they inherit many of the core challenges observed in ADAS deployments, particularly in detecting and responding to complex, multi-vehicle interactions in real-world scenarios~\citep{cafiso2012evaluation,yan2024comparison}. Both ADAS and ADS require robust sensor data, intelligent algorithms, and effective decision-making capabilities to ensure safety and reliability under diverse driving conditions~\citep{guo2019safe}.

Despite major technological strides, most widely deployed ACC and AEB systems—key features in both ADAS and early ADS—focus predominantly on the vehicle directly ahead. By doing so, they often fail to account for the behavior of trailing vehicles, leaving a significant gap in the safety envelope~\citep{nidamanuri2021progressive}. In real-world high-risk situations, such as dense, high-speed traffic, abrupt braking by a middle vehicle can lead to dangerous chain-reaction collisions if the following vehicles cannot decelerate in time. These pile-ups become even more severe if a heavy-duty vehicle is involved, given its longer stopping distance and greater momentum. Consequently, while current systems may prevent immediate rear-end collisions with the lead vehicle, they are far from optimal in managing more complex, multi-vehicle scenarios—particularly those that pose edge-case challenges for semi- or fully autonomous vehicles.

To address this shortfall in both ADAS and ADS contexts, this study proposes a novel longitudinal control and collision avoidance algorithm designed to integrate adaptive cruising with emergency braking. Our framework leverages deep reinforcement learning (DRL) to capture interactions with both preceding and following vehicles, thereby mitigating multi-vehicle collision risks. Unlike traditional approaches that primarily emphasize front-facing danger, this method broadens the control horizon to reduce chain-reaction events. Implemented via Deep Deterministic Policy Gradient (DDPG), our approach dynamically balances comfort and safety in various traffic conditions. Simulation results indicate that this DRL-based control strategy yields superior performance relative to conventional ACC and AEB~\citep{FR2023}, showcasing its potential for autonomous driving systems as well as enhanced ADAS.

A vital component of this solution is ensuring the reliability and realism of the training datasets. Sensor inaccuracies and noise~\citep{kang2020camera} can degrade the quality of raw data, while purely synthetic simulations often fail to reflect the complexities of real-world roads~\citep{luo2024survey}. These imperfections can impede proper exploration and convergence in reinforcement learning algorithms. To overcome these issues, we introduce a preprocessing framework that calibrates vehicle positional and trajectory data from camera feeds. By reducing measurement errors and biases, our solution fosters greater stability in virtual training environments, thereby improving the transferability of the learned policies to on-road scenarios for both ADAS and ADS applications.

The main contributions of this work are: 
\begin{itemize} \item \textbf{Longitudinal Control Policy for Leading and Following Vehicles:} Development of a unified braking and acceleration policy that proactively addresses collision risks from trailing vehicles by exploring edge-case scenarios, crucial for both ADAS and ADS.

\item \textbf{Novel Reward Function for Edge Cases:} Introduction of a reward function that incorporates potential collisions and inter-vehicle distances, enabling the DRL algorithm to handle hazardous, multi-vehicle conditions effectively.

\item \textbf{Universal Collision-Mitigation Algorithm:} A framework applicable to various autonomy levels, from semi-autonomous driving support to fully autonomous vehicles, thereby reducing the likelihood of severe pile-ups.

\item \textbf{Data Preprocessing Framework:} Proposal of a calibration strategy that refines raw sensor data to enhance simulation realism, diminish noise and errors, and improve reinforcement learning stability.

\item \textbf{Empirical Validation:} Demonstration through simulation studies that our DDPG-based approach significantly reduces collisions in complex traffic scenarios where conventional systems and naive ADS implementations are prone to failure.

\end{itemize}

In summary, by tackling the often-overlooked danger from following vehicles, this work lays a robust foundation for enhanced ADAS implementations and autonomous driving systems capable of handling high-risk, multi-vehicle situations. Through careful data calibration, an advanced DRL framework, and a focus on real-world edge cases, our approach offers a promising path toward safer, more resilient automotive technologies. The reminding of the paper are organized as follows: Section 2 provides a comprehensive review of the related studies in the existing literature; Section 3 provides a brief overview of the developed models; Section 4 presents the case study design and experiments of the proposed models; Section 5 offers the result analysis; and Section 6 summarizes the conclusions and key findings.

\section{Literature Review}

\subsection{Adaptive Cruise Control and Automatic Emergency Braking}
ACC and AEB have evolved as foundational features in both ADASs and ADSs, significantly contributing to vehicle autonomy. Initially designed to optimize fuel consumption and reduce emissions through predictive control strategies, ACC ensures a smooth driving experience by automatically adjusting speed and following distance based on real-time data gathered from onboard sensors~\citep{lu2019energy}. This adaptability enables ACC systems to handle diverse driving environments—from congested urban roads to high-speed highways—while maintaining safety and comfort~\citep{yu2022researches}.

Parallel to ACC, AEB utilizes sensor inputs and algorithms to detect imminent collisions and autonomously activate the braking system. By minimizing reaction times and maximizing braking efficiency, AEB reduces both the likelihood and severity of crashes~\citep{fildes2015effectiveness}. Critical to its effectiveness is the balance between precision and timeliness: the system must neither overreact (causing unnecessary braking events) nor underreact (risking collisions). Recent advancements in sensor fusion and object detection algorithms have bolstered AEB’s versatility, allowing it to address a wider range of hazardous situations—including interactions with pedestrians, cyclists, and other vulnerable road users.

Crucially, the integration of machine learning and artificial intelligence has accelerated the transition of ACC and AEB from passive support functions to active, predictive safety systems~\citep{moujahid2018machine}. By leveraging large-scale datasets and increasingly powerful computational techniques, these approaches enhance the precision and adaptability of control algorithms. They have also catalyzed innovation in traffic safety, facilitating the development of more complex ADAS/ADS features that can interpret real-time, high-dimensional sensor data and make split-second decisions~\citep{kim2017prediction}.

Despite these strides, existing ACC and AEB solutions typically emphasize only the vehicle directly ahead, neglecting critical interactions with trailing vehicles. This creates vulnerabilities in chain-reaction braking scenarios—particularly at high speeds—where abrupt deceleration by a single vehicle can lead to multi-vehicle collisions. As the automotive industry progresses toward higher levels of driving autonomy, overcoming this limitation becomes increasingly urgent to ensure robust safety and reliability in real-world traffic conditions.



\subsection{Reinforcement learning application in ADAS/ADS}
Reinforcement Learning (RL) has emerged as a promising paradigm for enhancing the decision-making capabilities of ADAS and ADS, enabling algorithms to learn optimal policies through continuous interaction with, and feedback from, their environment~\citep{wang2022deep}. Traditional rule-based and optimization methods often struggle in highly dynamic and uncertain driving scenarios, where the behavior of other vehicles, road conditions, and external factors can be unpredictable. RL circumvents these challenges by incrementally refining control actions—such as acceleration, braking, and steering—based on reward signals obtained during exploration.

Recent advances in computational power and the proliferation of robust simulation platforms—e.g., CARLA~\citep{dosovitskiy2017carla}—have significantly accelerated RL research in the automotive domain. Specifically, DRL algorithms like Deep Q-Networks (DQN) and DDPG have shown a heightened capacity to handle the state-action complexity of ADAS tasks~\citep{8441758}. By combining deep neural networks with RL frameworks, these algorithms learn high-level abstractions of sensor inputs and traffic conditions, enabling them to make more nuanced decisions under uncertainty.

This capability has been explored in various ADAS functions: lane-keeping assistance~\citep{sallab2016end}, ACC~\citep{desjardins2011cooperative}, and AEB~\citep{fu2020decision,chae2017autonomous}. In each of these applications, DRL-driven systems demonstrate improved adaptability to unpredictable behaviors of surrounding vehicles and other road agents~\citep{chen2023using, chen2024deep}. When integrated with sensor fusion—merging data from LiDAR, radar, camera, and other sensing modalities—DRL algorithms can build a richer understanding of the driving environment, thus making more accurate predictions and safer control decisions~\citep{9106866}.

Notably, the ability of RL to cope with partial observability and noisy sensor data is particularly relevant as vehicle autonomy grows~\citep{feng2023dense}. While higher levels of autonomy demand increasingly sophisticated perception and planning, RL-based controllers have the potential to incorporate learned experience from extensive simulated and real-world interactions—making them a powerful tool for advancing ADAS toward full ADS functionality~\citep{wang2024research}.

\subsection{Edge case scenario exploration}
Edge-case scenarios in both ADAS and ADS refer to rare or extreme driving conditions where conventional algorithms and sensor systems may fail or perform unreliably. These can include unusual weather conditions (e.g., heavy snow, thick fog), sudden and erratic pedestrian or cyclist movements, or non-standard vehicle maneuvers (e.g., abrupt lane changes or multi-vehicle chain reactions). Verifying vehicle performance and algorithmic robustness in such scenarios is crucial to guaranteeing safety, especially as the industry progresses toward higher levels of autonomy~\citep{koopman2016challenges}.

Research on modeling and simulating edge cases is extensive. Techniques often involve generating synthetic environments, stress-testing algorithms with adversarial conditions, or replaying real-world incidents under varied parameters~\citep{abbas2019safe,chen2025insight}. For example, closed-track testing focuses on pushing the vehicle and its control algorithms to the boundaries of their operational design domain, thereby uncovering weaknesses before widespread deployment~\citep{li2021risk}. Addressing these vulnerabilities early helps refine both the control policies and sensor fusion mechanisms, ensuring that edge-case failures are less likely to occur in real traffic.

By integrating comprehensive edge-case testing into the development lifecycle, engineers and researchers can proactively identify and correct model deficits. This is especially pertinent in chain-reaction collision scenarios, where the interactions among multiple vehicles are fluid and unpredictable. As such events can lead to catastrophic outcomes, robust handling of these edge cases becomes a linchpin for the safer deployment of ADAS and ADS technologies in everyday driving.

\begin{figure}
    \centering
    \includegraphics[width=1\linewidth]{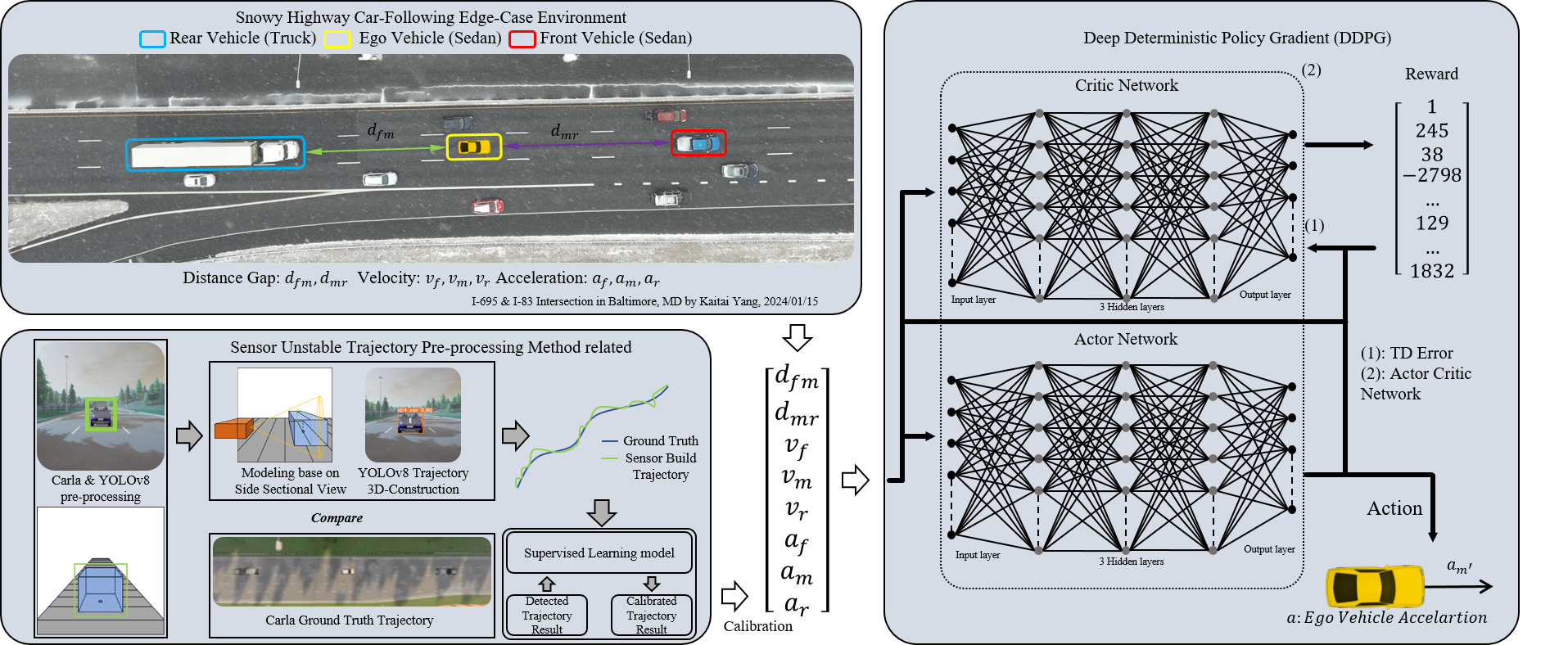}
    \caption{Model framework}
    \label{CV}
\end{figure}

\section{Methodology}

In this study, we develop a universally applicable deep reinforcement learning model, based on the Deep Deterministic Policy Gradient (DDPG) algorithm, that effectively handles vehicle-following scenarios on both highways and urban roads. This model manages complex car-following situations and accommodates various vehicle types and their physical and kinematic models. DDPG learns a Q-function and a policy simultaneously using off-policy data and actor-critic neural network architecture, employing the Bellman equation to update the Q-function, which guides policy learning. Additionally, our approach investigates edge cases typically elusive to optimization studies and integrates calibration of vehicle detection results from cameras, enhancing stability by minimizing error rates associated with estimating the positions of vehicles ahead and behind. This novel RL-based algorithm is particularly adept in longitudinal control and collision avoidance, effectively managing high-risk driving scenarios and accommodating different acceleration policies with enhanced precision.

\begin{figure}
    \centering
    \begin{subfigure}[b]{0.5\linewidth}
        \centering
        \includegraphics[width=\linewidth]{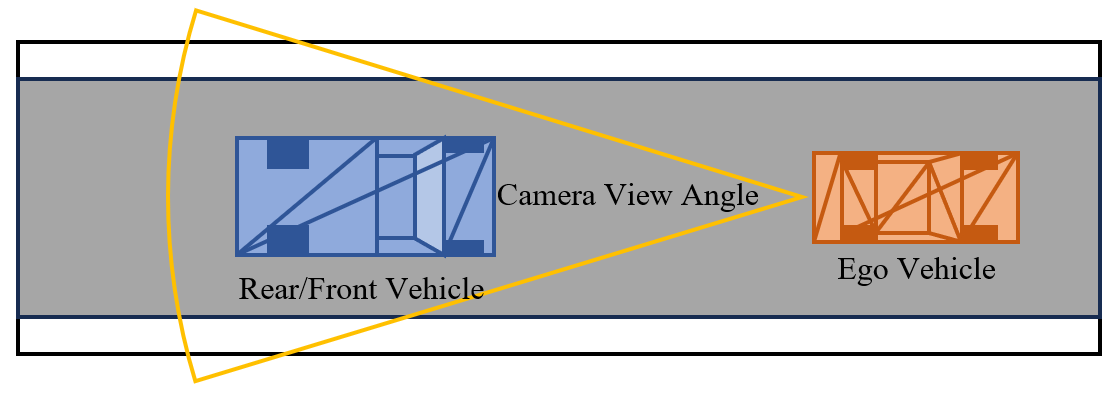}
        \label{fig:camera-view-model:(a)}
    \end{subfigure}

    \begin{subfigure}[b]{0.21\linewidth}
        \centering
        \includegraphics[width=\linewidth]{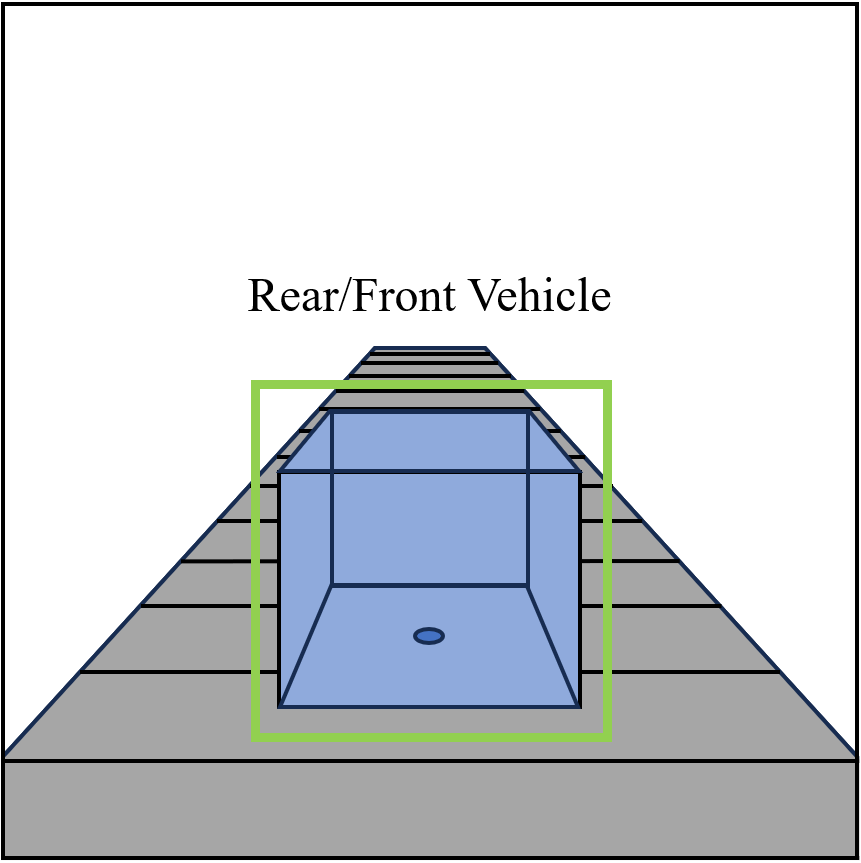}
        \label{fig:camera-view-model:(b)}
    \end{subfigure}
    \hspace{1mm}
    \begin{subfigure}[b]{0.27\linewidth}
        \centering
        \includegraphics[width=\linewidth]{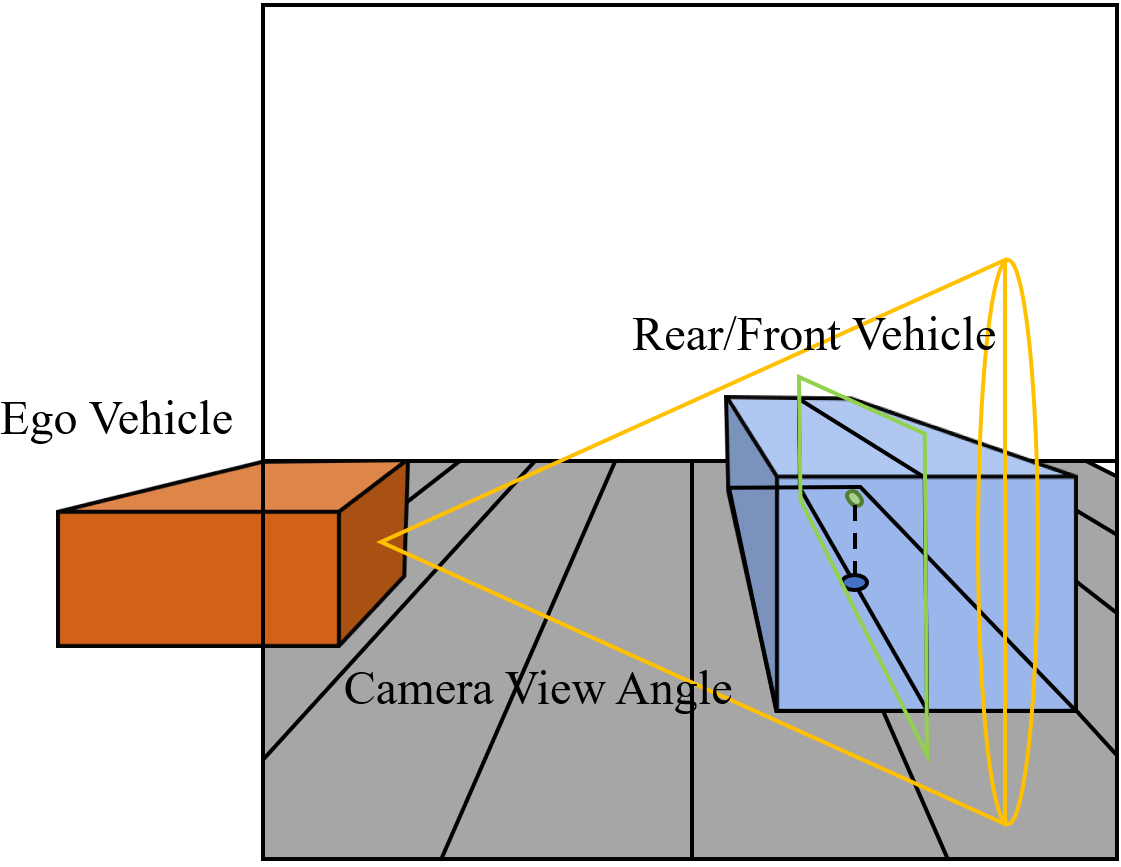}
        \label{fig:camera-view-model:(c)}
    \end{subfigure}
    \caption{Sensor perspective diagram.}
    \label{fig:camera-view-model}
\end{figure}

\subsection{Non-ego vehicle trajectory preprocessing model }

In the field of vehicle detection, a significant challenge is the uniform error of frame-by-frame detection accuracy. Most computer vision algorithms for this task provide a certain degree of accuracy but often overlook changes in the center point of the bounding box. This study uses the output of the CARLA simulator vehicle camera to process the bounding box trajectory of the trailing vehicle and reconstruct it in 3D from a top-down view.

A subtle issue exists when generating 3D vehicle models from sensor data, affecting multi-sensor integration and camera-only detection. The core problem is the deviation of the center of the reconstructed vehicle from its actual position. To solve these inaccuracies, especially in remote vehicle detection scenarios, a novel vehicle detection model is introduced. By comparing real vehicle trajectories and detection results in CARLA, the YOLOv8 algorithm for rear camera vehicle detection is established along with an error distribution model. This model refines the output of camera detection and incorporates input from the DDPG environment to improve vehicle localization accuracy, aiming to provide a more accurate representation of vehicle location.

Integrating the CARLA simulation and video recognition systems to pre-process sensor outputs is crucial for calibrating the reinforcement learning (RL) state. The importance of this integration is highlighted by the observation that smaller vehicles in the image (indicating greater distances from the camera sensor) significantly affect the accuracy and center positioning of the detection bounding box due to variance and error recognition. To address these issues, a preprocessing method has been developed that includes acquiring ground truth data for CARLA vehicles, collecting datasets from CARLA's ego vehicle rear camera, and deploying a supervised learning model. This model refines the calibration between the ground truth and the 3D trajectory constructed from the sensor detection results.

To acquire the position of the leading or following vehicle using sensor data (Front/Rear Camera), the preprocessing model was divided into several parts:
\subsubsection{Camera Perspective Transformation}
The proposed model needs to process the camera's image and transform the vehicle detection trajectory within one vertical plane to another horizon plane (top-view plane). This transformation simulation will maintain the linearity of lines in the image but alter their relative size and shape to eliminate the perspective effect where distant objects appear smaller than those closer. 

\begin{figure}
  \centering
  \includegraphics[scale=0.25]{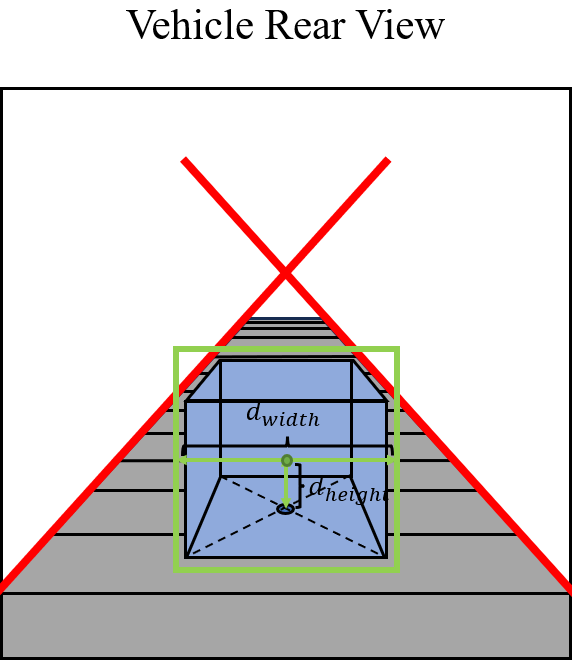}
  \hspace{0.1in}
  \includegraphics[scale=0.25]{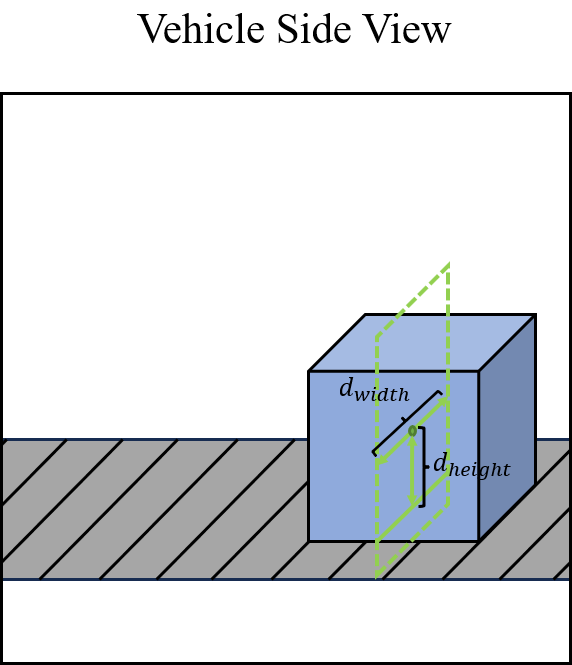}
  \caption{Vehicle center localization}
  \label{fig:localizationmethod}
\end{figure}

A brief and general localization model is proposed to localize the vehicle's center point which is shown in Fig.~\ref{fig:localizationmethod}. This model is suitable for camera sensors and combines the features of the different vehicle sizes. With the YOLOv8 detection bounding box, the bounding box width $d_{width}$, height $d_{height}$,and center point $(x_b,y_b)$ can be acquired. The average sedan size is about 12 feet by 5.9 feet by 4.7 feet. The projection $(x_p,y_p)$ of the vehicle center on the ground is:
\begin{equation}
    x_p = x_b, y_p= y_b-\frac{h_{vehicle}\times d_{width}}{w_{vehicle}}    
\end{equation}

Mathematically, the perspective transformation matrix can be represented by a $3\times3 $ matrix. The general form of the perspective transformation is:
\begin{equation}
    \begin{pmatrix}
    x' \\
    y' \\
    w'
    \end{pmatrix}
    =
    \begin{pmatrix}
    a & b & c \\
    d & e & f \\
    g & h & i
    \end{pmatrix}
    \begin{pmatrix}
    x \\
    y \\
    1
    \end{pmatrix}
\end{equation}
where $(x,y)$ is a coordinate set of the original image point, $(x',y')$ is the coordinate set after transformation, $w'$ is the third coordinate which used to maintain the perspective effect in the homogeneous coordinate system. The final coordinates $(x',y')$ are normalized by dividing by $w'$:
\begin{equation}
    x_{final} = \frac{x'}{w'}, y_{final} = \frac{y'}{w'}
\end{equation}

\begin{figure}[!t]
        \centerline{\includegraphics[width=0.35\textwidth]{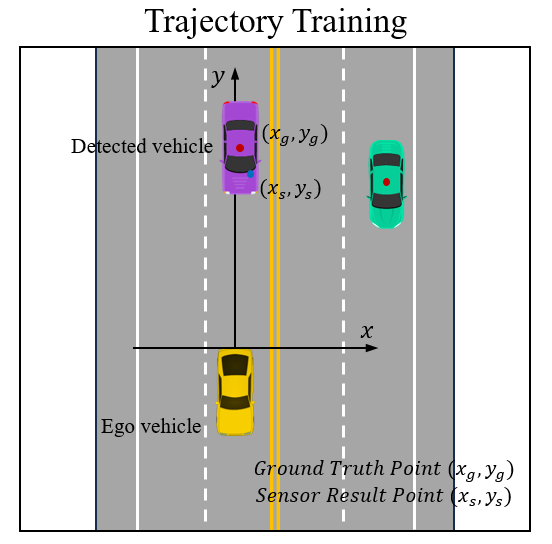}}
        \caption{Ground Truth Points Set and Sensor Result Points Set}
        \label{fig:Topview_2}
    \end{figure}

The perspective transformation matrix for the application in image correction of autonomous vehicles needs to be calculated. Four points located in the straight car lane for the direct rear camera are taken, and the trapezoid is transformed into a top-view rectangle. The transformation points set is shown in Fig.~\ref{fig:Topview_2}.
\begin{figure}[!t]
        \centerline{\includegraphics[width=0.5\textwidth]{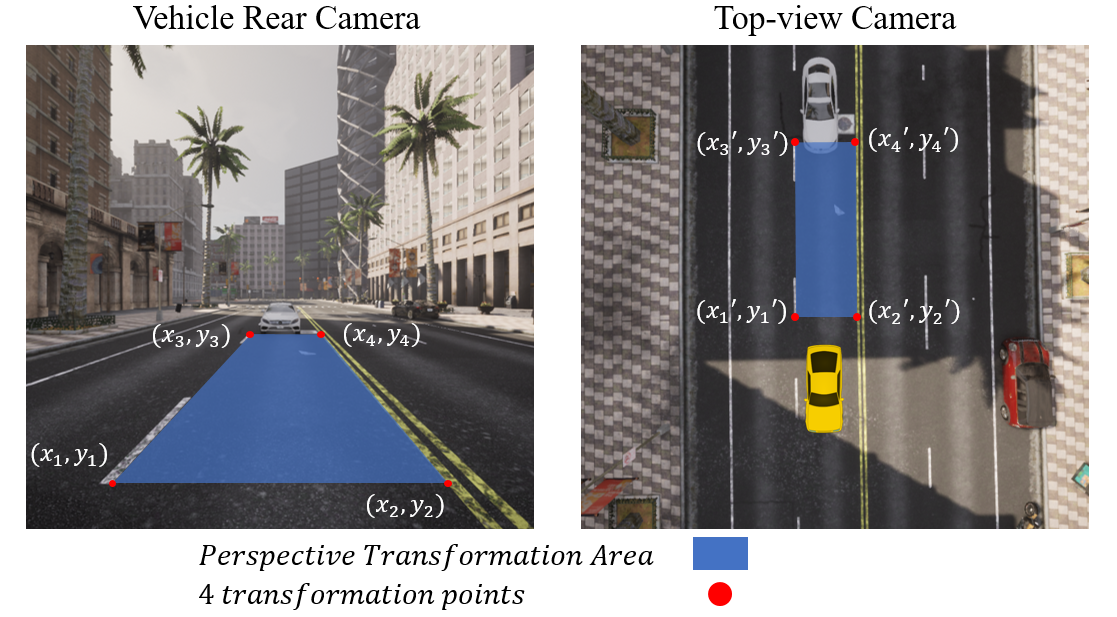}}
        \caption{Perspective Transformation Area and Points Set}
        \label{fig:Topview_1}
    \end{figure}

Here we used the computer vision library OpenCV to calculate the matrix $\begin{pmatrix}
 a & b & c \\
    d & e & f \\
    g & h & i
    \end{pmatrix},$ 
The non-zero scaling factor $w=1$ is normally chosen to simplify the homogeneous coordinate $(x',y',w')$ to the Cartesian coordinate $(x'/w',y'/w')$ and to normalize the coordinate to the final 2-D coordinate. In this case, the new resulting coordinates are in the form $(x_i', y_i', w_i')$, where $w_i'$ is the new $w$ component and corresponds to the expression $gx_i + hy_i + i$. The actual equations to solve are:
\begin{equation}
    x'_i = \frac{ax_i + by_i + c}{gx_i + hy_i + 1}, \quad y'_i = \frac{dx_i + ey_i + f}{gx_i + hy_i + 1}
\end{equation}

In case to solve the matrix elements, we rearrange the equations to a linear system by multiplying both sides by the denominator to eliminate the division. The final matrix of the result linear equations are:
\begin{equation}
   \begin{bmatrix}
   x_1 & y_1 & 1 & 0 & 0 & 0 & -x'_1x_1 & -x'_1y_1 \\
    0 & 0 & 0 & x_1 & y_1 & 1 & -y'_1x_1 & -y'_1y_1 \\
    x_2 & y_2 & 1 & 0 & 0 & 0 & -x'_2x_2 & -x'_2y_2 \\
    0 & 0 & 0 & x_2 & y_2 & 1 & -y'_2x_2 & -y'_2y_2 \\
    \vdots & \vdots & \vdots & \vdots & \vdots & \vdots & \vdots & \vdots \\
    x_4 & y_4 & 1 & 0 & 0 & 0 & -x'_4x_4 & -x'_4y_4 \\
    0 & 0 & 0 & x_4 & y_4 & 1 & -y'_4x_4 & -y'_4y_4 \\
    \end{bmatrix}
    \begin{bmatrix}
    a \\
    b \\
    c \\
    d \\
    e \\
    f \\
    g \\
    h \\
    \end{bmatrix}
    =
    \begin{bmatrix}
    x'_1 \\
    y'_1 \\
    x'_2 \\
    y'_2 \\
    \vdots \\
    x'_4 \\
    y'_4 \\
    \end{bmatrix}
\end{equation}

\subsection{Markov decision process}
The ego vehicle agent used an environment state with 8 parameters, an action, and a well-designed reward function. The complicated car following scenario can be represented by a Markov Decision Process (MDP) with the tuple $(S,A,P_a,r_a)$
\subsubsection{State Space (S)}
A set of states. $S$ represents each time step's environment state variables when we consider the RL vehicle agent as the middle ego vehicle. $S$ includes the distance between the leading vehicle and ego vehicle $d_{fm}$, distance between the ego vehicle and following vehicle $d_{mr}$, leading vehicle velocity $v_f$, ego vehicle velocity $v_m$, following vehicle velocity $v_r$, leading vehicle acceleration $a_f$, ego vehicle acceleration $a_m$, following vehicle acceleration $a_r$. The ego vehicle's sensor is only able to capture the current position of itself and the distance between itself and the leading/following vehicles. A Kalman Filter is used to estimate the velocity and acceleration.

\subsubsection{Action Space (A)} 
A set of actions. $A$ represent the next time step's acceleration $a_{m'}$ of each episodes' time step. The ego vehicle can get the next time step's acceleration $a_{m'}$ to interact with the environment and update the State $S$.
\subsubsection{Transition Probability (P)}
The transition probability is:
\begin{equation}
    P_a(s,s') = Pr(s_{t+1}=s'|s_t=s,a_t=a)
\end{equation} 
$P_a(s,s')$ is the probability of changing from state $s$ to next time step state $s'$ when take the ego vehicle acceleration action $a_{m'}$.
\subsubsection{Reward Function (r)}
A set of rewards. $r$ represent the reward function $r(s_{t+1},s_t,a_t)$. $r$ is the expected immediate reward of taking a specific action $a$ from state $s$ to state $s'$.

We are trying to find the policy function $\pi(s)$ that is able to generate the optimal action for a particular state to  maximize the expectation of cumulative future rewards:
\begin{equation}
\label{cum_reward}
    E\left[\sum_{t=0}^{\infty} \gamma^t r\left(s_t, s_{t+1},a_t\right)\right]
\end{equation}
where $\gamma$ is the discount factor and the range of $\gamma$ is $[0,1]$. The larger $\gamma$  motivates the RL agent to favor taking actions indefinitely, rather than postponing them early.

\subsection{Distance and Collision based reward function}
\subsubsection{Reward function design}
\begin{figure}
        \centerline{\includegraphics[width=0.45\textwidth]{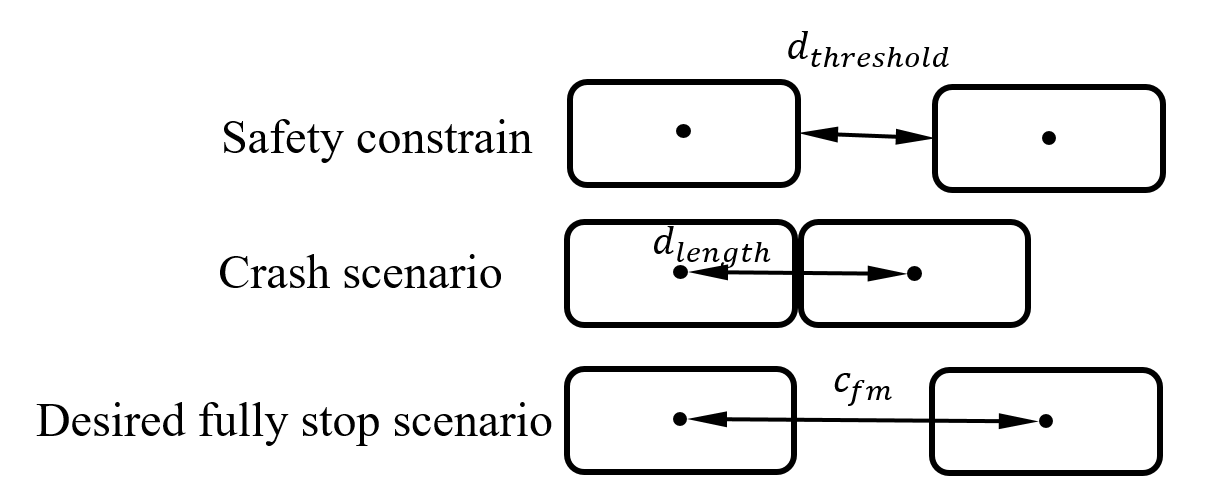}}
        \caption{State transition based on distance and collision}
        \label{fig:fullystoptemp}
    \end{figure}

In our proposed DRL model, there exist two state transitions which are shown in Figure
\begin{itemize}
    \item Crash happens between the middle ego vehicle and the leading/following vehicle. The reward function for the ego RL vehicle agent at this time step's transition is $C_0$ which is usually designed as a large negative number, like $-3000$.
    \item Ego RL vehicle successfully followed the leading vehicle and did not crash with both the leading and following vehicle. The reward of this time-step transition is:
    \begin{equation}
        C_3-(\frac{f_{fm}}{d_{fm}-c_{fm}+C_1}+\frac{f_{mr}}{d_{mr}-c_{mr}+C_2})
    \end{equation}
    where $f_{fm}$ and $f_{mr}$ are the important factors to distinguish whether the algorithm about takes more consideration to the front-middle or middle-rear vehicle crash, $c_{fm}$ and $c_{mr}$ is the sum of vehicle length and the desired minimum gap distance threshold for the physical environment, and $C_1$, $C_2$,$C_3$ is the constant to avoid singularity. The desired fully stop scenario is shown in Fig.~\ref{fig:fullystoptemp}
    
\end{itemize}

The cumulative reward obtained by the RL agent is another critical metric, reflecting the agent's performance in terms of safety and efficiency. It is calculated by summing the immediate rewards $r_t$ at each time step $t$ over the simulation horizon $T$:

\begin{equation} 
\label{r_total}
R_{\text{total}} = \sum_{t=1}^{T} \gamma^{t} r_t
\end{equation}

where the $\gamma \in [0,1]$ is the discount factor, the $T$ represents the end step of each training episode, and each time step reward $r_k$ is defined in Eq.~\ref{reward_func1}
\begin{equation}
    r_t = \begin{cases}
        C_0, &\text{If collision in $t^{th}$ time step}\\
        \displaystyle C_3-(\frac{f_{fm}}{d_{fm}-c_{fm}+C_1}+\frac{f_{mr}}{d_{mr}-c_{mr}+C_2}), & \text{If no collision in $t^{th}$ time step}
    \end{cases}
    \label{reward_func1}
\end{equation}

Thus the combination of the accumulated reward $R_t$ is:
\begin{equation}
    R_t = \sum_{t=1}^{T} \gamma^{t} \begin{cases}
        C_0, &\text{If collision in $t^{th}$ time step}\\
        \displaystyle C_3-(\frac{f_{fm}}{d_{fm}-c_{fm}+C_1}+\frac{f_{mr}}{d_{mr}-c_{mr}+C_2}), & \text{If no collision in $t^{th}$ time step}
    \end{cases}
\end{equation}

This reward structure encourages the agent to learn policies that minimize the risk of collision while maintaining efficient traffic flow.

\subsubsection{Convergence derivation}
A detailed formula derivation is proposed below to prove the designed reward function can create the optimal equilibrium state:\\
Assume the distance between the leading and following vehicle is constant $D$, $c_{fm}$ and $c_{mr}$ is also constant.

The $d_{fm}$ and $d_{mr}$ satisfy:
\begin{equation}
    d_{fm}+d_{mr}=D, d_{fm}=x, d_{mr}=D-x
\end{equation}

Given these definitions, the reward function for non-collision scenarios is simplified to:
\begin{equation}
    r_t(x)=C_3-(\frac{f_{fm}}{x-c_{fm}+C_1}+\frac{f_{mr}}{D-x-c_{mr}+C_2})
\end{equation}

In this expression, $C_3$ acts as a baseline reward. The two subtracted terms are penalties that depend inversely on the “effective distances” (with adjustments $c_{fm}$ and $c_{mr}$, and offsets $C_1$ and $C_2$). Essentially, if the distances deviate too far from a desired configuration, the penalties increase, reducing the reward.
To simplify the algebra, we define two new functions:
\begin{equation}
    A(x) = x-c_{fm}+C_1, B(x) = D-x-c_{mr}+C_2
\end{equation}

With these definitions, the reward function can be rewritten as:
\begin{equation}
    r_t(x)=C_3-[\frac{f_{fm}}{A(x)}+\frac{f_{mr}}{B(x)}]
\end{equation}

Since $C_3$ is a constant, maximizing $r_t(x)$ is equivalent to minimizing the penalty term inside the brackets. For convenience, we define a function $f_t$ as follows:
\begin{equation}
    f_{t}(x) = \frac{f_{fm}}{A(x)}+\frac{f_{mr}}{B(x)}
\end{equation}

To main rule on the first item is to ensure the vehicle can maintain its position between the leading and following vehicles and that the maximum DDPG algorithm's reward will not cause a collision in the current timestep.
\begin{equation}
    \frac{f_{fm}}{A(x)}=f_{fm} \cdot A(x)^{-1}
\end{equation}

The derivative with respect to $x$ is:
\begin{equation}
    \frac{d}{dx} \left[ f_{fm} \cdot A(x)^{-1} \right] = f_{fm} \cdot \frac{d}{dx}\left[A(x)^{-1}\right]
\end{equation}
\begin{equation}
\label{de1item2}
\frac{dA(x)}{dx} = \frac{d}{dx}\left[x - c_{fm} + C_1\right] = 1
\end{equation}

According to the power function derivation formula, we can derive $A(x)^{-1}$ based on Eq.~\ref{de1item2} to get:
\begin{equation}
\begin{aligned}
\label{de1item}
    \frac{d}{dx}\left[A(x)^{-1}\right] &= -1 \cdot A(x)^{-2} \cdot \frac{dA(x)}{dx}, \\
    &= -A(x)^{-2}    
\end{aligned}
\end{equation}
The negative sign here is crucial and reflects that the reciprocal function decreases as its denominator increases.

Thus, the derivative of the first term is:
\begin{equation}
\frac{d}{dx} \left[ \frac{f_{fm}}{A(x)} \right] = -\frac{f_{fm}}{A(x)^2} = -\frac{f_{fm}}{(x - c_{fm} + C_1)^2}
\label{diff_1}
\end{equation}
This result quantifies how sensitive the penalty from the front vehicle is to changes in $x$.

Next, we consider the second term. Its derivative is computed similarly:
\begin{equation}
    \frac{d}{dx}\left[ \frac{f_{mr}}{B(x)} \right] = \frac{f_{mr}}{B(x)^{-2}} = \frac{f_{mr}}{(D - x - c_{mr} + C_2)^2}
    \label{diff_2}
\end{equation}

Combining these results gives the total derivative of $f_t(x)$:
\begin{equation}
    f_t'(x) = -\frac{f_{fm}}{(x - c_{fm} + C_1)^2} + \frac{f_{mr}}{(D - x - c_{mr} + C_2)^2}
\end{equation}
This derivative indicates the net rate of change of the penalty as $x$ varies. Setting it equal to zero will yield the equilibrium condition.
To find the optimal equilibrium point $x^*$, assume the $R'(x) = 0$, then can get:
\begin{equation}
   \frac{f_{mr}}{(D-x-c_mr+C_2)^2}=\frac{f_{fm}}{(x-c_{fm}+C_1)^2} 
\end{equation}
This equality ensures that any increase in one penalty is exactly offset by a decrease in the other, indicating a balance.

Taking the square root of both sides (which is allowed since the expressions are positive by design) gives:
\begin{equation}
    \frac{\sqrt{f_{mr}}}{D-x-c_{mr}+C_2}=\frac{\sqrt{f_{fm}}}{x-c_{fm}+C_1}
\end{equation}

We now introduce a simplifying parameter:
\begin{equation}
    \beta = \sqrt{\frac{f_{mr}}{f_{fm}}}
\end{equation}
This parameter $\beta$ captures the relative scaling between the penalties associated with the rear and front vehicles.

Using $\beta$, we can write:
\begin{equation}
    \begin{aligned}
        x-c_{fm}+C_1 &=\beta(D-x-c_{mr}+C_2)\\
        &=\beta D-\beta x -\beta c_{mr}+\beta C_2
    \end{aligned}
\end{equation}
This relation shows that the effective distance from the front (after adjustment) is proportional to the effective distance from the rear.

Finally, solving for $x$ yields the equilibrium point:
\begin{equation}
    x^*=\frac{\beta (D-c_{mr}+C_2)+c_{fm}-C_1}{1+\beta}
    \label{x*}
\end{equation}
This expression represents the unique state at which the reward function attains its global maximum, ensuring that the vehicle is positioned optimally between the front and rear vehicles.

To verify that $x^*$ indeed corresponds to a maximum of the reward function (or equivalently a minimum of the penalty function), we compute the second derivative. For the first term and second item:
\begin{equation}
\begin{aligned}
    \frac{d^2}{dx^2}(\frac{f_{fm}}{x-c_{fm}+C_1})&=\frac{2f_{fm}}{(x-c_{fm}+C_1)^3}\\
    \frac{d^2}{dx^2}(\frac{f_{mr}}{D-x-c_{mr}+C_2})&=\frac{2f_{fm}}{(D-x-c_{mr}+C_2)^3}
\end{aligned}
\end{equation}
Both second derivatives are positive because the denominators are cubed positive quantities (by the design constraints).

Finally, we can acquire:
\begin{equation}
    f''_t(x)= \frac{2f_{fm}}{(x-c_{fm}+C_1)^3}+\frac{2f_{fm}}{(D-x-c_{mr}+C_2)^3}
\end{equation}

During the design, we ensure that the denominators are all positive, so $f''_t(x)>0$ holds for all valid $x$, that is, $r_t(x)$ is a convex function and its global minimum point is $x^*$ from Eq.~\ref{x*}. Because $f_t(x)$ will reach the global minimum at $x^*$, $r_t(x)$ can acquire the global maximum at $x^*$, meeting the accumulated reward maximization target.

\subsubsection{Stability analysis}
 In the current DDPG framework, the actor neural network outputs the action acceleration $a$, the vehicle dynamic function will influence the state $s$ which is specifically related to the distance $x$ of the leading and following vehicle.

 The whole process can be described as:
 \begin{itemize}
 \item If there exists a difference $\Delta x$ between the current $x$ and the optimal $x^*$, the DDPG algorithm will output acceleration $a$ for the next time step to drive the system states to $x^*$.
 \item When $x$ is converged to the optimal location $x^*$, ideally the acceleration $a$ and velocity $v$ will approach the optimal solution $a^*$ and $v^*$ for maintaining the vehicle can stay in optimal $x^*$.
 \item Due to the reward function $r_t(x)$ paper adopted reaching the maximum $x^*$, the DDPG algorithm will lead the policy state $s(x)$ to converge  $s(x^*)$ during the accumulating the maximum return.

 \end{itemize}

\textbf{Policy gradient derivation: }To discuss the  policy gradient and draw out the stabilization of system, the per-time-step reward in Eq.~\ref{reward_func1}
with the cumulative (discounted) reward defined by Eq.~\ref{r_total} will be used as first step (The collision reward $C_0$ is a constant and will not contribute to the gradient update).
Let the policy be deterministic, denoted by \(\mu_\theta(s)\), where \(\theta\) represents the parameters of the Actor network and \(s\) is the state (which includes \(x\)). The goal is to maximize the expected cumulative reward:
\begin{equation}
J(\theta) = \mathbb{E}_{s\sim \rho^\mu}\left[Q\big(s, \mu_\theta(s)\big)\right]
\end{equation}

where the Q-function is defined in Eq.~\ref{cum_reward}.

According to the Deterministic Policy Gradient (DPG) theorem, the gradient of \(J(\theta)\) is given by:
\begin{equation}
\nabla_\theta J(\theta) = \mathbb{E}_{s\sim \rho^\mu}\left[ \nabla_\theta \mu_\theta(s) \cdot \nabla_a Q(s,a)\Big|_{a=\mu_\theta(s)} \right]
\end{equation}

Due to the specific expression of $r_t$ is a piecewise function, when there is no collision happening, the vehicle state $s$ include $x$, and the action $a$ will influence the $x$ by vehicle dynamic rules:
\begin{equation}
    x_{t+1}=f(x_t,a_t)
\end{equation}

Then the action $a$ affects the subsequent reward $r(x_t)$ by affecting $x_t$, which makes the reward function have an implicit dependency on the action. Next step is to compute the gradient of \(Q(s,a)\) with respect to the action \(a\). Note that \(a\) influences the state \(x\), which in turn affects the reward. By the chain rule:
\begin{equation}
\begin{aligned}
&\nabla_a Q(s,a) = \sum_{t=1}^{T} \gamma^t \, \nabla_a r(x_t) =\nabla_a r(s)+\gamma \nabla_a Q(s',a')\\
&where \space \nabla_a r(s)= \nabla_x r(x) \cdot \nabla_a x
\end{aligned}
\end{equation}

For each time step \(t\),
\begin{equation}
\nabla_a r(x_t) = \frac{\partial r(x_t)}{\partial x_t} \cdot \frac{\partial x_t}{\partial a}
\end{equation}

Next, we compute \(\frac{\partial r(x)}{\partial x}\) from the reward function Eq.~\ref{reward_func1}
And also can get the differentiate term from Eq.~\ref{diff_1} and Eq.~\ref{diff_2}. 
Assuming the state dynamics are differentiable with respect to the action \(a\):
\begin{equation}
\frac{\partial x_t}{\partial a} \quad \text{as the sensitivity of the state \(x_t\) to the action \(a\)}
\end{equation}

Then, the gradient of the Q-function with respect to \(a\) becomes:
\begin{equation}
\nabla_a Q(s,a) = \sum_{t=1}^{T} \gamma^t \left( \frac{f_{fm}}{(x_t - c_{fm} + C_1)^2} - \frac{f_{mr}}{(D - x_t - c_{mr} + C_2)^2} \right) \cdot \frac{\partial x_t}{\partial a}
\end{equation}

Substitute the computed \(\nabla_a Q(s,a)\) into the deterministic policy gradient expression:
\begin{equation}
\nabla_\theta J(\theta) = \mathbb{E}_{s\sim \rho^\mu}\left[ \nabla_\theta \mu_\theta(s) \cdot \left(\sum_{t=1}^{T} \gamma^t \left( \frac{f_{fm}}{(x_t - c_{fm} + C_1)^2} - \frac{f_{mr}}{(D - x_t - c_{mr} + C_2)^2} \right) \cdot \frac{\partial x_t}{\partial a} \right) \Bigg|_{a=\mu_\theta(s)} \right]
\end{equation}

Finally, the actor network updates its parameters based on this gradient, so that its output action $a$ can drive the system state $s$ to gradually converge to the optimal point $x^*$, maximize the long-term cumulative reward, and stabilize the RL agent in proposed complicated driving behaviors.

\subsection{Evaluation metrics}
In our methodology, several evaluation metrics have been used to assess the performance of the RL agent controlling the middle vehicle in a three-vehicle platoon. The primary metric is the 'collision rate', which measures the safety of the RL-controlled vehicle under varying acceleration conditions of the leading and following vehicles.

To evaluate the collision rate of the proposed RL algorithm, we simulate a three-vehicle platoon where both the leading and following vehicles have a deceleration range of $[-7.5 m/s^2,0 m/s^2]$
 ] during the deceleration phase. A collision will be considered as occurring if the gap between the ego (middle) vehicle and either the leading or following vehicle is smaller than the threshold of 2 meters. If a collision is detected, the algorithm is not considered a safe solution under these conditions.

The collision rate evaluation is calculated as:

\begin{equation}
\text{Collision Rate} = \frac{\text{Total Number of Collisions}}{\text{Total Number of Simulations}}
\end{equation}

This metric provides an overall measure of safety across all simulation runs.

Specifically focusing on the RL-controlled middle vehicle, we calculate the ego middle vehicle collision probability:

\begin{equation} \label{eq:probcollision}
P_{\text{collision}} = \frac{N_{\text{Middle Vehicle Collisions}}}{N_{\text{Total Possible Non-Collision Cases}}}
\end{equation}

This metric assesses the likelihood of the middle vehicle being involved in a collision when collisions are not considered unavoidable by the baseline model. $N_{\text{Total Possible Non-Collision Cases}}$ means the possibility of if there is enough space for the ego middle vehicle between the leading and following vehicle in the whole deceleration process. The evaluate space is defined as the gap distance of the leading and following vehicles which will start deceleration at the same beginning time step of the simulation and ignore the ego middle RL vehicle.

By analyzing these metrics across various acceleration scenarios for the front and rear vehicles, it's possible to evaluate the RL agent's ability to adapt its driving strategy in dynamic environments.

\section{Experiment}
\subsection{Baseline model}
A baseline model combining AEB and ACC is developed to evaluate the proposed algorithm. The Time-To-Collision (TTC) is used as the AEB and ACC engagement factor. The vehicles' velocities are kept at the same to reduce the perturbations~\citep{8884686}. If the TTC between the ego vehicle and the leading vehicle is less than $1.4s$~\citep{donnell2009speed} which is calculated from the Federal Highway Administration emergency stop's deceleration concept $0.7-0.85g$, the ego vehicle will activate the emergency brake system. The algorithm of the baseline ADAS model is shown in Algorithm~\ref{ADASalgorithm}.
\begin{algorithm}[!ht]
	\caption{Baseline ADAS Algorithm} 
	\label{ADASalgorithm} 
	\begin{algorithmic}[1]
        \REQUIRE Leading vehicle related position $(x_f,y_f)$, ego  vehicle position $(x_{m},y_{m})$, leading vehicle initial velocity $v_{fi}$, ego vehicle velocity $v_{m}$, time step $\Delta t$
		\STATE Initialize Variables
        \STATE Start Episode
        \FOR{$i=1$ \TO $Maximum$ $Episode$}
        \FOR{$n=1$ \TO $Episode$ $Steps$}
        \STATE Predict leading vehicle velocity $v_{f}$ by Kalman Filter
        \STATE Predict leading vehicle new position $(x_{fnew},y_{fnew}) = (x_{f},y_{f}) + v_{f} * \Delta t$
        \IF{$TTC_{front to middle}<1.4s$}
        \STATE Active ego vehicle deceleration $a_{ego}=-7.5m/s^{2}$ \citep{donnell2009speed}
        \ENDIF
        \ENDFOR
        \ENDFOR
		
        \end{algorithmic} 
\end{algorithm}

\subsection{Sensor data preprocessing model}
To preprocess the data from the sensor and eliminate the variance and error recognition for specific models, a method that combined CARLA and YOLOv8 was developed. After enabling the CARLA traffic manager function and generating a dense urban-highway comprehensive driving scenario, a camera sensor was set up at the rear end of one vehicle. The scenario was run for 30 minutes, generating a continuous 30-minute rear camera video dataset. Clips that had the following vehicle in the same ego vehicle car lane and could be successfully recognized by the sensor were manually selected.

Due to different sensors and the sensor fusion combination, the interior trajectory generation method varies. Thus, the calibration machine learning model was used to explore each unique vehicle detection and localization algorithm. After the trajectory was acquired by perspective transformation, the post-transform trajectory and the ground truth data set were classified into the training dataset and validation dataset. The neural network consists of 5 fully connected layers, including an input layer, an output layer, and three hidden layers. The hyperparameters of the FCNN are shown in Table~\ref{hyperparameterfcnn}.
\begin{table}
    \centering
    \begin{tabular}{c|c }
         Hyperparameter & Value \\
         \hline
         Input and Output& $(x',y')$ and $(x_{cali}',y_{cali}')$\\
         \hline
         Neural Network Size & $[2, 1024,512,256,2]$ \\
         \hline
         Hidden Layers Number & $2$ \\
         \hline
         Activation Function & ReLU, ReLU, ReLU, None\\
         \hline
         Learning Rate & $0.001$ \\
         \hline
         Replay Buffer Batch Size & $512$ \\
         \hline
         Regularization & $L_2$ Regularization\\
         \hline
         Weight decay & $0.0001$\\
         \hline
         Loss function & MSE \\
         \hline
         Optimization algorithm & Adam \\
         
    \end{tabular}
    \caption{Hyperparameters for FCNN }
    \label{hyperparameterfcnn}
\end{table}

\subsection{Proposed RL methodology}
Two reinforcement methodologies for ego middle vehicle agents, which have different reward function designs, were proposed. To satisfy the exploration and exploitation in the DDPG algorithm, different scenarios were combined to simulate the vehicle's following scenario during model training:
\begin{itemize}
    \item The leading and following vehicle agents maintain their velocity base a Gaussian Distribution. The ego middle vehicle agent gains its velocity by using the acceleration output from the actor-critic neural network. 
    \item With a trigger that also has a Gaussian Distribution to simulate the emergency brake in different scenarios, the leading vehicle will active deceleration with a higher mean Gaussian Distribution and the following vehicle will active deceleration with a lower mean Gaussian Distribution when the TTC time to the middle ego vehicle is smaller than $1.4s$.
\end{itemize}

\begin{algorithm}[!t]
	\caption{Proposed RL-based ADAS Algorithm} 
	\label{DDPGalgorithm} 
	\begin{algorithmic}[1]
        \REQUIRE Current time step vehicle sensor detection result $data_f$ and $ data_r$, Ego vehicle position $(x_{m},y_{m})$, Ego vehicle velocity $v_m$
        \STATE Start episode
        \FOR{$i = 1$ \TO $Maximum$ $Episode$}
        \FOR{$n = 1$ \TO $Episode$ $Steps$}
        \STATE Acquire the middle ego RL vehicle's position $(x_m,y_m)$, velocity $v_m$ and the approximate gap distance $d_{fm}$ $d_{mr}$ between Leading/following vehicle and ego middle vehicle from $data_f$ and $data_r$ 
        \STATE Estimate next time step related vehicle position $(x_f,y_f)$, $(x_r,y_r)$, velocity $v_f$, $v_r$, acceleration $a_f$, $a_r$
        
        \IF{Collision happened}
        \STATE Return collision reward
        \STATE End current training episode
        \ELSE
        \STATE Return the successful following reward which is precisely designed.
        \STATE Interact with the environment. Turn to the current episode's next time step. Return to $Step (3)$
        \ENDIF

        \ENDFOR
        \ENDFOR

	\end{algorithmic} 
\end{algorithm}

\subsection{Edge case scenario}
\begin{figure}[t!]
    \centering
    \begin{subfigure}[b]{0.49\textwidth}
        \centering
        \includegraphics[width=\linewidth]{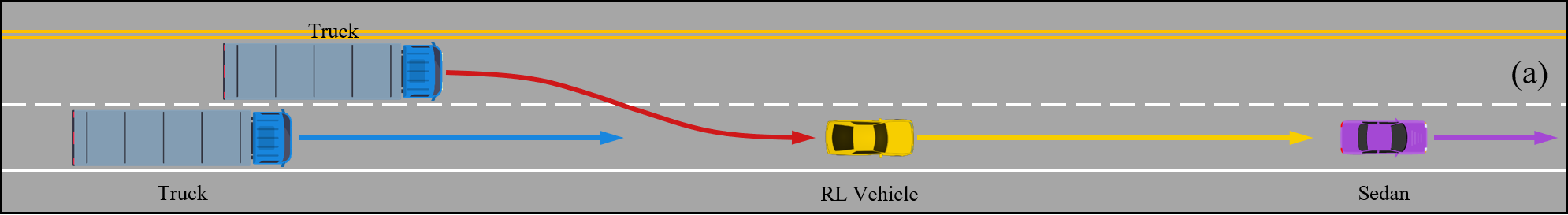}
        \caption{}
        \label{fig:sc1}
    \end{subfigure}
    \hspace{1mm}
    \begin{subfigure}[b]{0.49\textwidth}
        \centering
        \includegraphics[width=\linewidth]{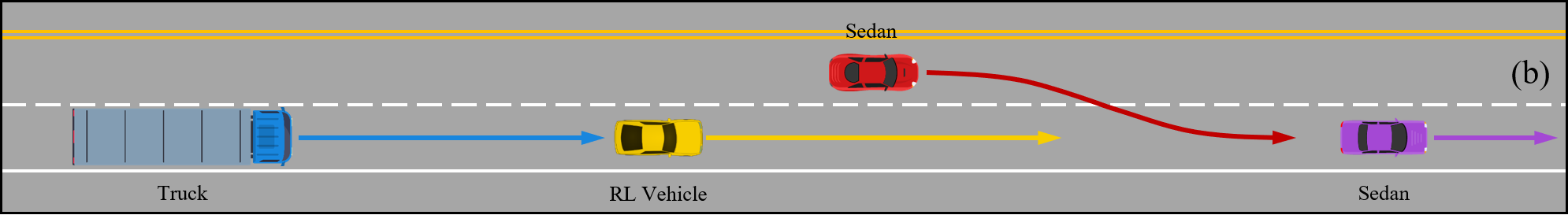}
        \caption{}
        \label{fig:sc2}
    \end{subfigure}
    
    \begin{subfigure}[b]{0.49\textwidth}
        \centering
        \includegraphics[width=\linewidth]{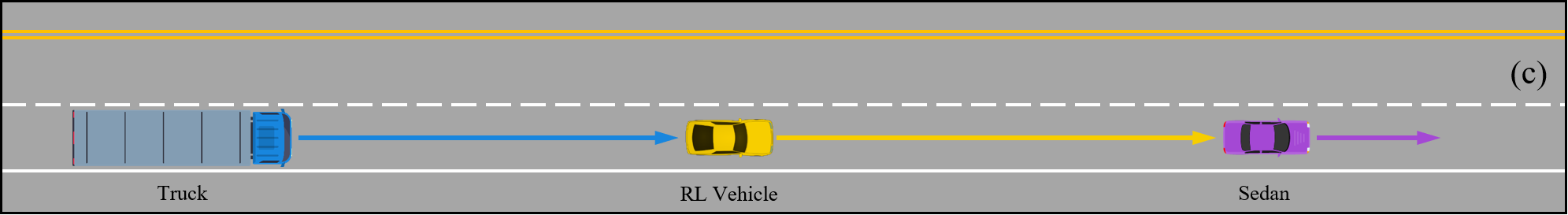}
        \caption{}
        \label{fig:sc3}
    \end{subfigure}
    \hspace{1mm}
    \begin{subfigure}[b]{0.49\textwidth}
        \centering
        \includegraphics[width=\linewidth]{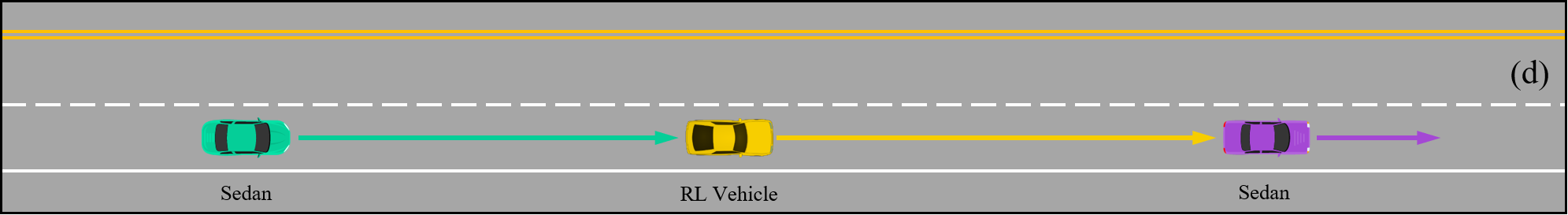}
        \caption{}
        \label{fig:sc4}
    \end{subfigure}
    
    \begin{subfigure}[b]{0.49\textwidth}
        \centering
        \includegraphics[width=\linewidth]{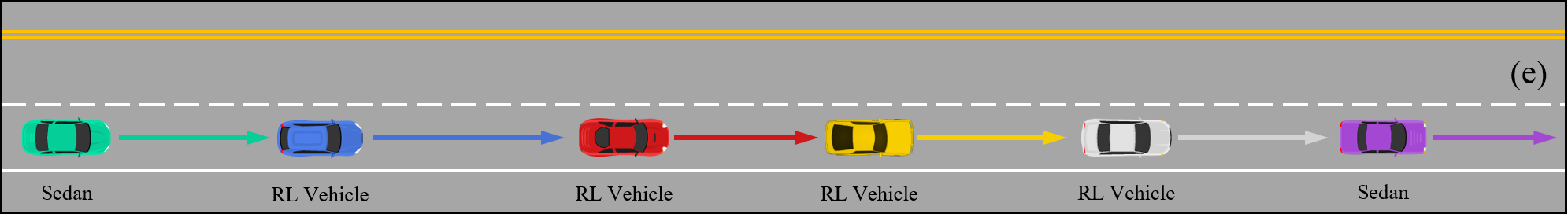}
        \caption{}
        \label{fig:sc5}
    \end{subfigure}
    \hspace{1mm}
    \begin{subfigure}[b]{0.49\textwidth}
        \centering
        \includegraphics[width=\linewidth]{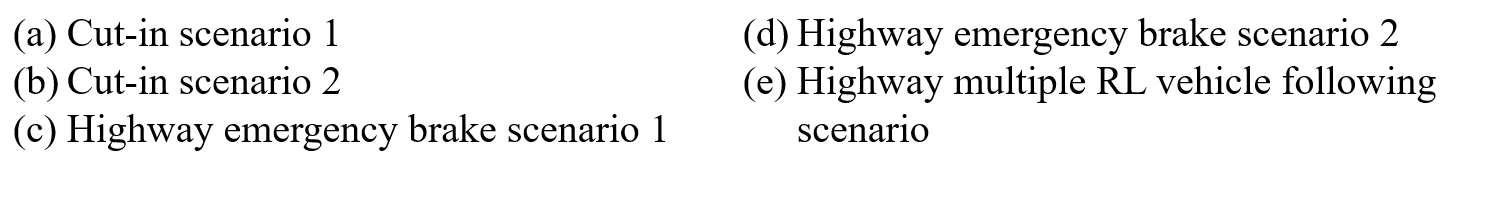}
        \caption{}
        \label{fig:scconclude}
    \end{subfigure}
    \caption{Potential dangerous collision scenarios.}
    \vspace{-0.1in}
    \label{fig:collision_scenarios}
\end{figure}

For many RL cases, it's hard to check whether the designed scenario is already capable for the algorithm to converge to an effective solution. Thus, we designed several typical scenarios that make it easy to cause the collision not only with the leading vehicle but also with the following vehicle to verify the algorithm's capability to deal with the dilemma that the current baseline ADAS driving algorithm can not solve. The five scenarios include cut-in scenarios, highway emergency brake scenarios, and the highway multiple RL vehicles following scenario. The scenario sketch diagram is shown in Fig.~\ref{fig:collision_scenarios}.


\begin{table}[!t]
    \abovetopsep = 0pt
    \aboverulesep = 0pt
    \belowrulesep = 0pt
    \belowbottomsep = 0pt
    \centering
    \begin{tabular}{c|c|c}
        \toprule
         Edge case Scenario & Key point & Vehicle Size\\
         \midrule
         \multirow{3}*{Cut-in scenario}&  \multirow{3}*{Leading vehicle cut-in}& Leading vehicle: Light \\
         ~ & ~ &  Cut-in vehicle: Light \\
         ~ & ~ & Following vehicle : Heavy \\
         \hline
         \multirow{3}*{Cut-in scenario}& \multirow{3}*{Following vehicle cut-in} & Leading vehicle: Light \\
         ~ & ~ & Cut-in vehicle: Heavy \\
         ~ & ~ & Following vehicle: Heavy\\
         \hline
         \multirow{1}*{Highway emergency}& \multirow{1}*{Leading vehicle } & Leading vehicle: Light\\
          brake scenario & emergency brake & Following vehicle: Heavy\\
         \hline
         \multirow{1}*{Highway emergency }& \multirow{1}*{Leading vehicle } & Leading vehicle: Light\\
         brake scenario & emergency brake & Following vehicle: Light\\
         \hline
         \multirow{3}*{\shortstack{Multiple RL \\vehicles following \\scenario }}& \multirow{3}*{RL agent capability} & \multirow{3}*{\shortstack{Leading vehicle: Light \\ Following vehicle: Heavy}}\\
         ~ & ~ & ~\\
         ~ & ~ & ~\\
         \hline

    \end{tabular}
    \caption{Proposed potential dangerous collision scenarios}
    \label{dangerousscenario}
\end{table}
\subsection{Implementation}
Compared to other RL methods used in crash avoidance and velocity maintaining, the basic and majority purpose of our ego RL vehicle agent is making sure the agent can avoid the collision with both leading and following vehicles and is capable of rare edge case situations. It's also important to make sure the Neural Net can learn from the final collision scenarios. Thus, the main point of the scenario design is to ensure during the exploration procedure, the collision scenario will take a significant portion of the whole training and can be used for the vehicle agent to maximize the reward and make the vehicle safer. 

Also, in the preprocessing model, the main point is to acquire a sensor (Rear vehicle camera) output data to compare the ground truth trajectory data and then get a supervised learning model to calibrate the future data before training the RL model. Between current methodologies, the CARLA simulator or real-world autonomous vehicle's sensor and GPS localization can help us acquire both sensor dataset and ground truth trajectory. To simplify the experiment's complexity and focus more on the RL algorithm, we chose the CARLA simulator to acquire the ground truth trajectory data.

The simulation is combined with the CARLA simulator, YOLOv8, Python, Stable-baseline3, and Gym. We proposed the training and vehicle method in Python 3.8.10 based on the RL implementation framework that is Stable-Baseline3 2.1.0 and gym 0.26.2 and based on the vehicle detection framework that is Ultralytics YOLOv8 8.1.6. We modified the RL environment based on the Gym environment "Pendulum-v1".
\paragraph{Scenario Parameters}
The vehicle positions are $(X, Y)$ and $Y=0m$, the initial position $X_f \sim N(36,0.5)m$, $X_m \sim N(18,0.5)m$, and $X_r \sim N(0,0.5)m$, the brake deceleration of the leading vehicle is $a_{f} \sim N(-7.5,0.2)m/s^2$, the leading vehicle stop starts at $t_f \sim U(1,1.5)s$, and the each time step's acceleration for leading and following vehicle is $a \sim N(0,0.01)m/s^2$. These normal-distributed vehicle parameters help the agent to explore most rare scenarios.

\paragraph{DDPG architecture and hyperparameters}
As shown in Fig.~\ref{CV}, we establish four fully connected networks as the actor-critic neural network. Each network has three hidden layers and uses the array \\$[State \, dimension, 256,256,256, Action\, dimension]$ as the network node. In the DDPG algorithm, the hyperparameters are shown in Tab.~\ref{hyperparameter}.

\begin{table}
    \abovetopsep = 0pt
    \aboverulesep = 0pt
    \belowrulesep = 0pt
    \belowbottomsep = 0pt
    \centering
    \begin{tabular}{c|c }
        \toprule
        Hyperparameter & Value \\
        \midrule
        \multirow{2}*{Neural Network Size} & $[State \, dim, 256,$ \\
        ~ & $256,256,Action\,dim]$\\
        \hline
        Hidden Layers Number & $3$ \\
        \hline
        Soft Update factor & $0.005$ \\
        \hline
        Memory Capacity & $10000$ \\
        \hline
        Replay Buffer Batch Size & $512$ \\
        \hline
        Discount Factor & $0.99999$ \\
        \hline
        Actor Network Learning Rate & $0.001$ \\
        \hline
        Critic Network Learning Rate & $0.002$ \\
        \hline

    \end{tabular}
    \caption{Hyperparameters for DDPG }
    \label{hyperparameter}
\end{table}

\section{Result}
Firstly, the DDPG reinforcement learning methodology was implemented, followed by the implementation of the Baseline ACC and AEB simulation model in the proposed potential dangerous scenarios shown in Table~\ref{dangerousscenario}. After that, the trajectory calibration model was pre-trained and the reinforcement learning methodology was tested in the scenarios outlined in Table~\ref{dangerousscenario} scenarios.

\subsection{DDPG training result}

\begin{figure}[!t]
        \centerline{\includegraphics[width=0.5\textwidth]{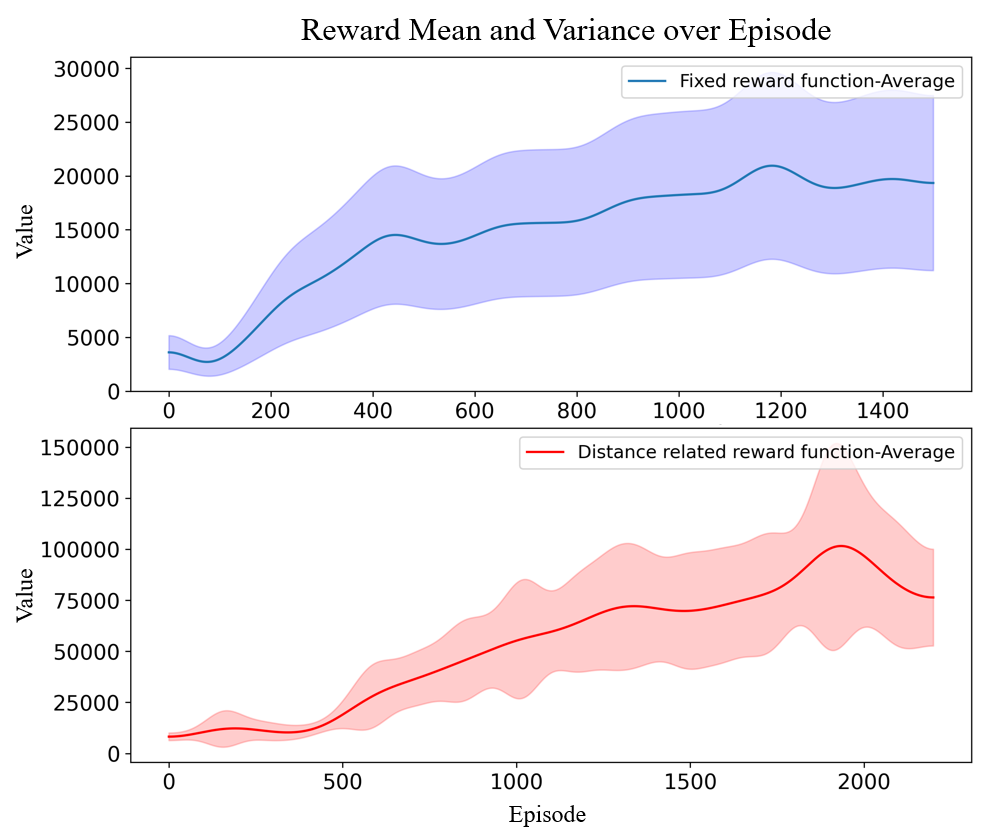}}
        \caption{Proposed RL algorithm Reward function}
        \label{fixedreward}
\end{figure}

The DDPG agent was trained in the proposed environment. A Python file named $modify\_random\_seed.py$ was created to execute the $trainDDPG.py$ script for training the reinforcement learning agent. The random seed set was $[20,50]$, which included 30 different random seeds. At the beginning of the training process, the training episode was set to $10000$ to observe if the neural net was over-fitting and to assess the training result. The two different training scripts (fixed reward and distance reward) were well-tuned with reward episodes of $1500$ and $2500$ that can significantly avoid collision, respectively. The final training process of the fixed reward scenario converged around the $400^{th}$ episode, with the reward, if the vehicle could fully avoid a collision with any of the vehicles, being $22500$. This training process was verified with 30 random seeds. Due to the different random seeds, the convergence reward function also varied. The dark line and light shading represent the average reward function and standard deviation. The reward function is shown in Fig.~\ref{fixedreward}. The final training process of the distance related to each time step reward scenario converged around the $1000^{th}$ episode, with the reward varying based on the distance the vehicle chose to maintain. This training process was also verified with 30 random seeds. The reward function is shown in Fig.~\ref{fixedreward}.

\subsection{Baseline ADAS simulation result}
The proposed ADAS driving algorithm is implemented in designed scenarios 1-4. Except for the leading vehicle, which has an emergency brake activated at a pre-defined time step, the other vehicles all follow a TTC collision threshold and will emergency brake at the maximum deceleration.

\begin{figure*}[h!]
    \centering
    \begin{subfigure}[t]{0.32\textwidth}
        \centering
        \includegraphics[width=\textwidth]{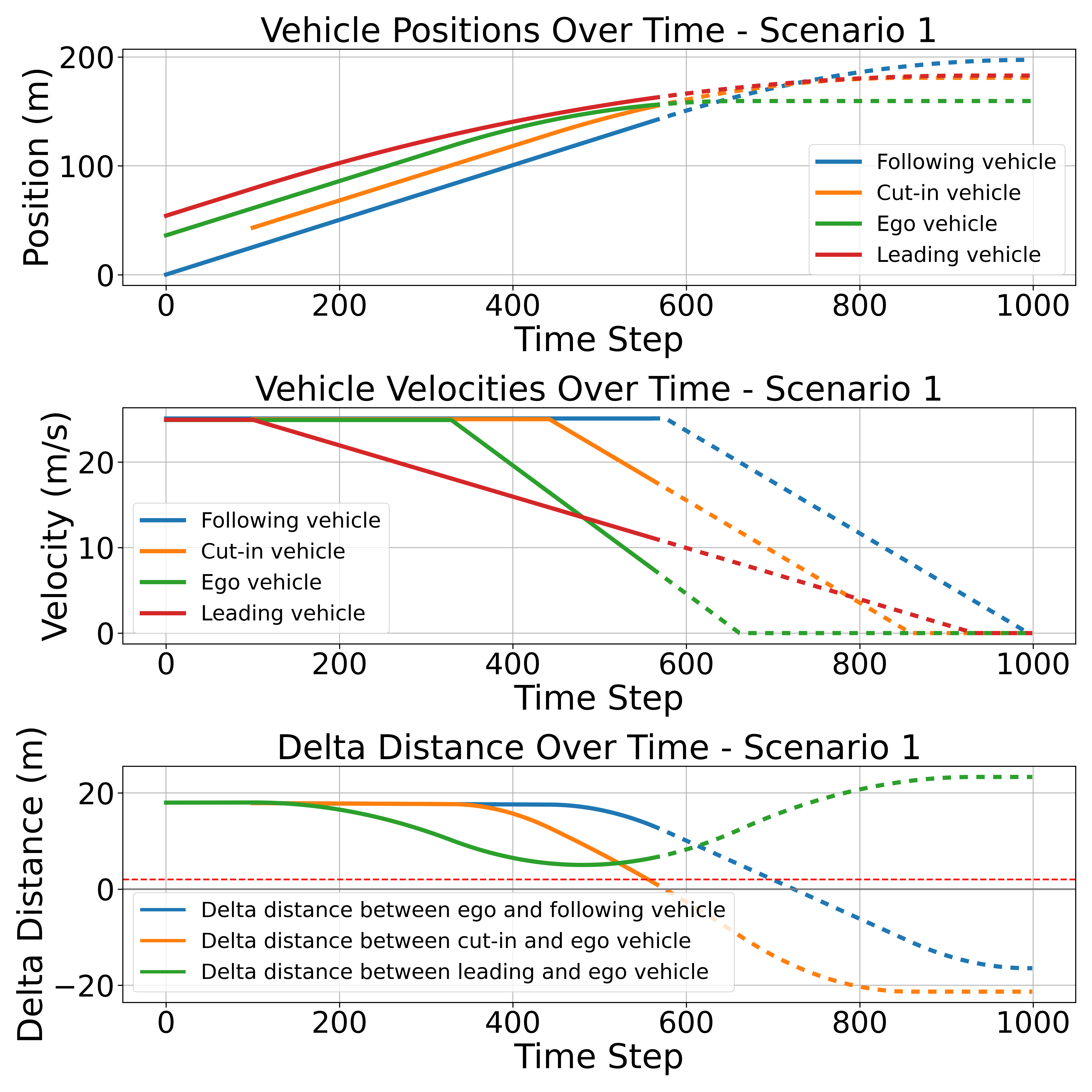}
        \caption{Cut-in scenario 1}
    \end{subfigure}
    \hspace{0mm}  
    \begin{subfigure}[t]{0.32\textwidth}
        \centering
        \includegraphics[width=\textwidth]{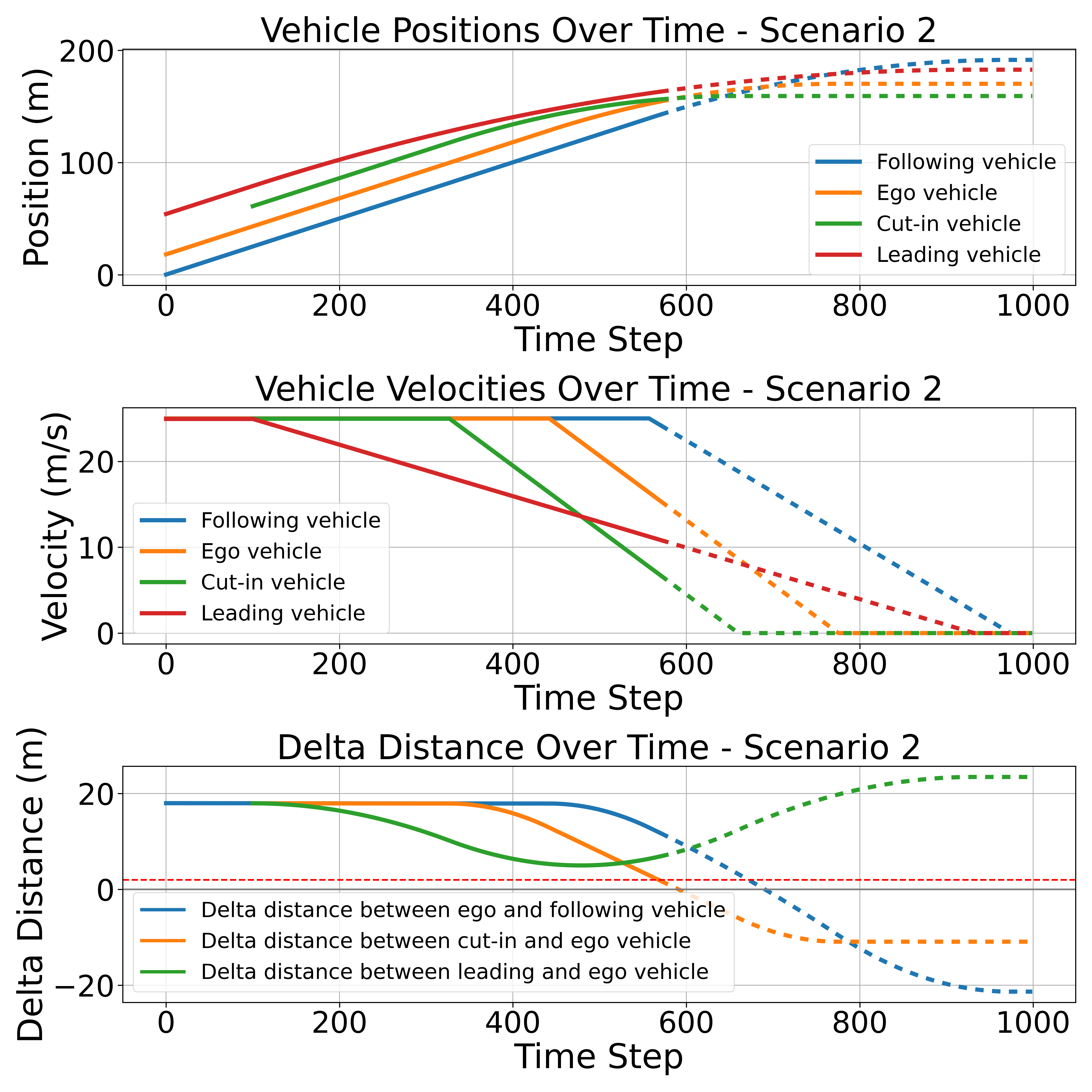}
        \caption{Cut-in scenario 2}
    \end{subfigure}
    \hspace{0mm} 
    \begin{subfigure}[t]{0.32\textwidth}
        \centering
        \includegraphics[width=\textwidth]{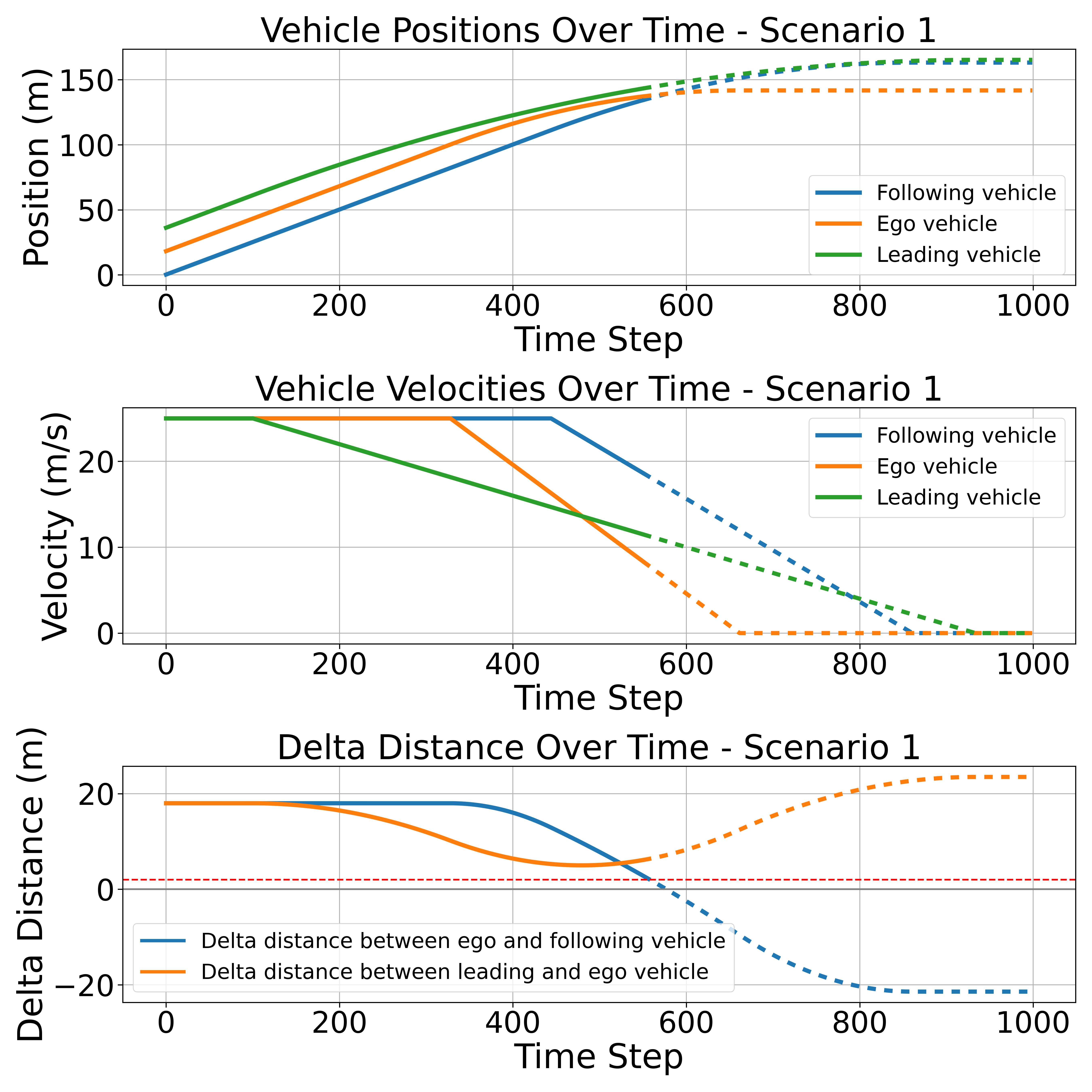}
        \caption{Highway emergency brake scenario 1}
    \end{subfigure}
    \quad
    \begin{subfigure}[t]{0.32\textwidth}
        \centering
        \includegraphics[width=\textwidth]{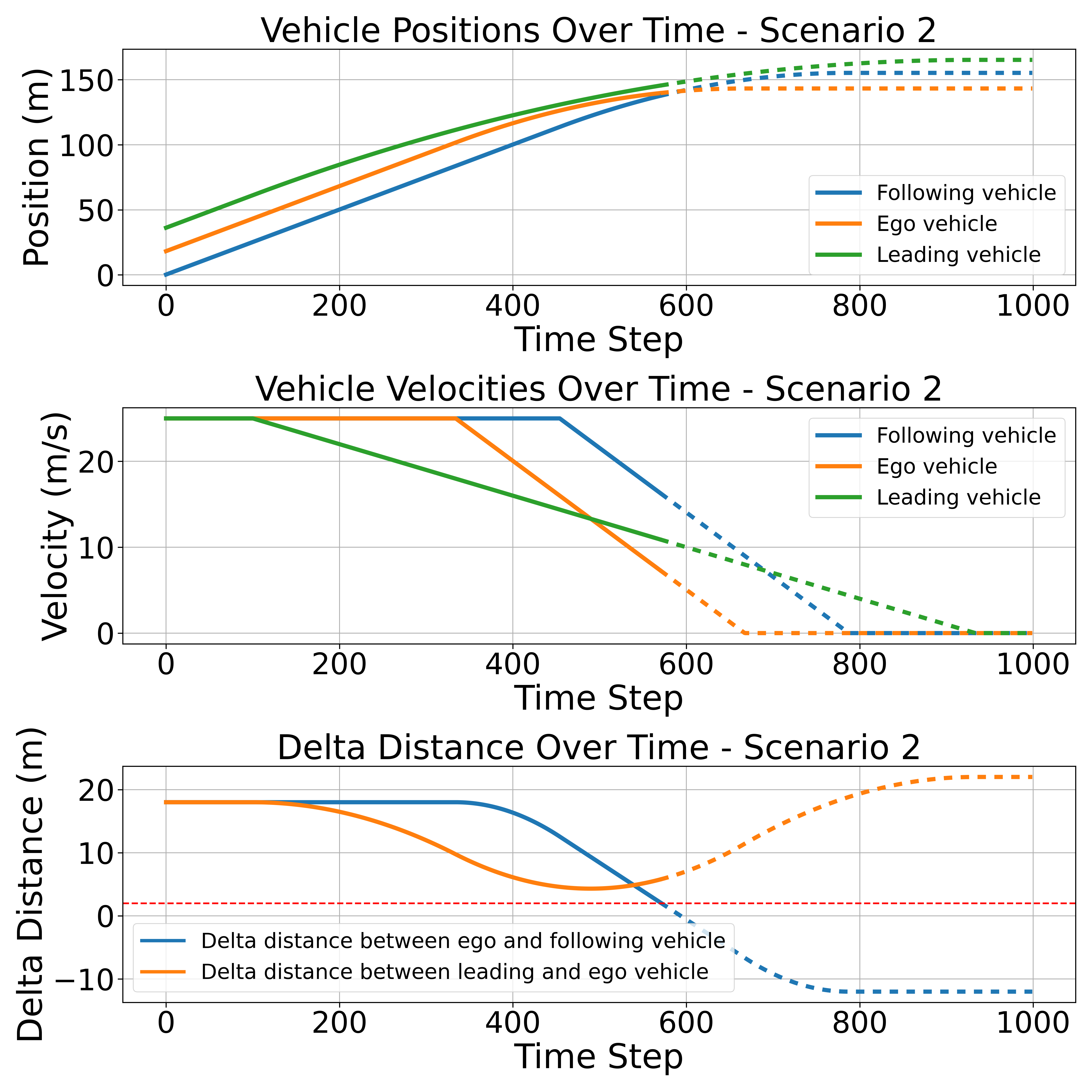}
        \caption{Highway emergency brake scenario 2}
    \end{subfigure}
    \hspace{0mm} 
    \begin{subfigure}[t]{0.32\textwidth}
        \centering
        \includegraphics[width=\textwidth]{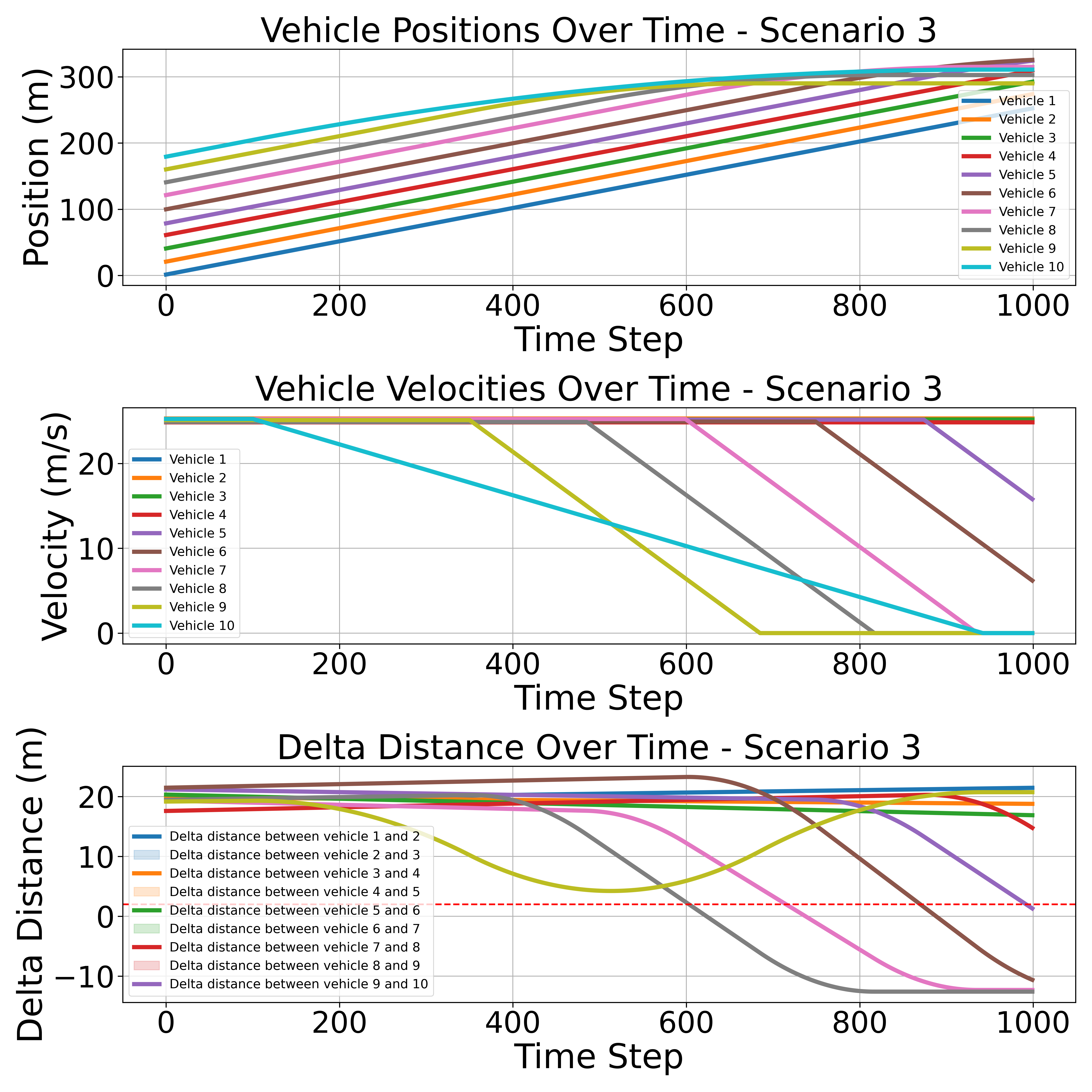}
        \caption{Multi-vehicle emergency brake scenario}
    \end{subfigure}
    \caption{Baseline ADAS algorithm implementation in proposed edge case scenarios}
    \label{Vehicleposition}
\end{figure*}

\subsubsection{Cut-in scenario result}
ADAS algorithm implementation in scenarios 1 and 2 are shown in Fig.~\ref{Vehicleposition}. In scenario 1, the yellow vehicle (truck) cuts in at 100 time step, and after the vehicle cuts in, the leading vehicle start deceleration. After a few time steps, each vehicle's AEB system also acts. Because of the smaller deceleration of the cut-in yellow vehicle (truck), the cut-in vehicle caused a collision with the ego vehicle which applied the ADAS driving algorithm. The delta distance between the ego vehicle and cut-in vehicle minimum distance is less than 2m which represents that the collision happens.

In scenario 2, the green vehicle (sedan) cuts in at $100^{th}$ time step, and after the vehicle cuts in, the leading vehicle starts deceleration. After a few time steps, each vehicle's AEB system also acts. Because of the same deceleration of the cut-in green vehicle (sedan) and less gap distance, the cut-in vehicle caused a collision with the blue following vehicle (truck) which applied the ADAS driving algorithm but with a smaller maximum deceleration. The delta distance between the ego vehicle and the following vehicle minimum distance is less than 2m representing that the collision happens.

\subsubsection{Highway emergency brake scenario}
ADAS algorithm implementation in scenarios 3 and 4 are also shown in Fig.~\ref{Vehicleposition}. We designed the blue following vehicle as a truck in scenario 3 and a sedan in scenario 4. The truck has a smaller maximum deceleration compared to the sedan in our designed emergency brake scenario. Thus, the collision will happen between the blue following vehicle (truck) in scenario 3 only due to the different deceleration features of different kinds of vehicles.
\subsection{Proposed RL algorithm simulation result}
After the RL algorithm is successfully converged, the proposed distance-related RL algorithm is implemented in the 5 proposed edge case scenarios.
\begin{figure}[!ht]
    \centering
    \begin{subfigure}[t]{0.32\textwidth}
        \centering
        \includegraphics[width=\textwidth]{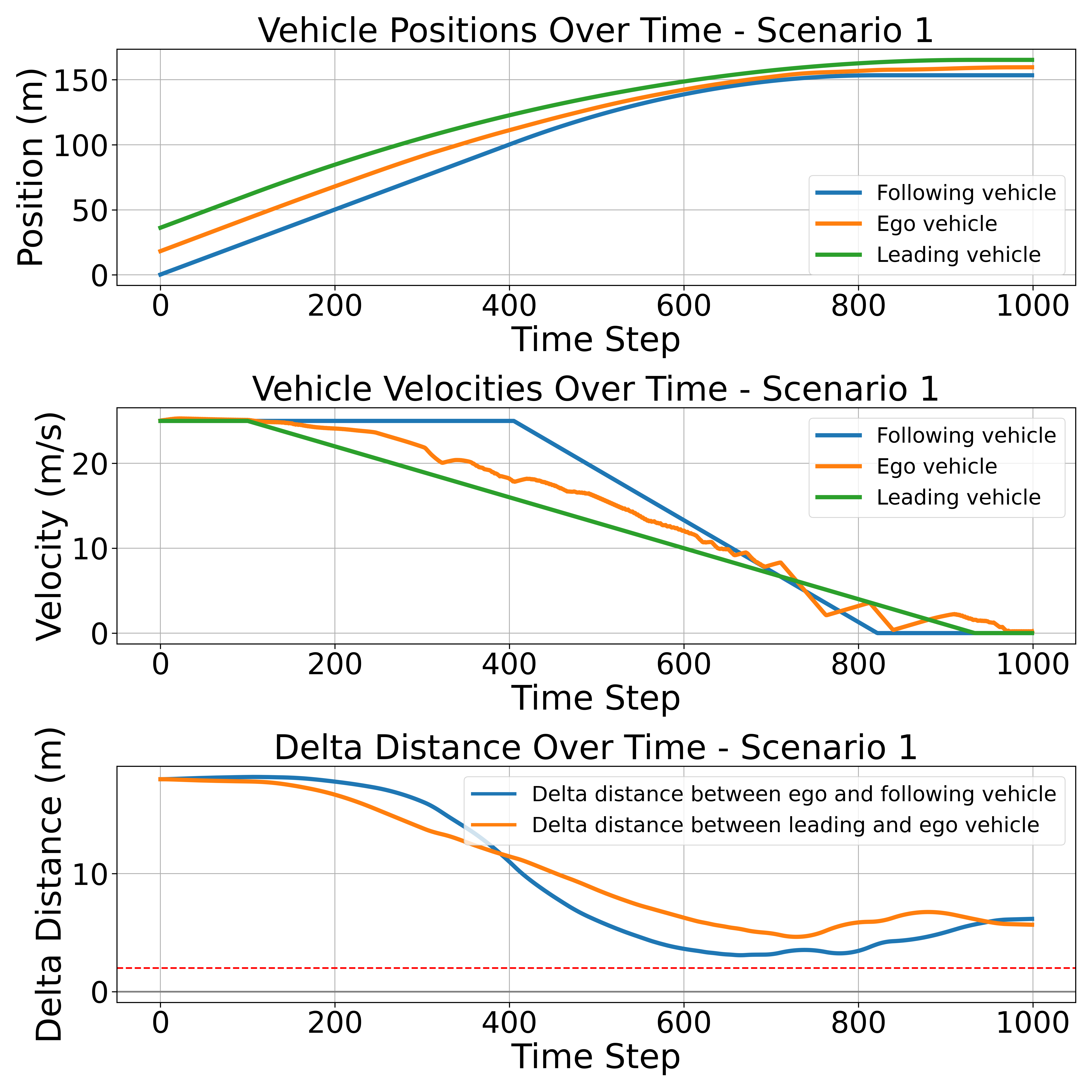}
        \caption{Cut-in scenario 1}
    \end{subfigure}
    \hspace{0mm}  
    \begin{subfigure}[t]{0.32\textwidth}
        \centering
        \includegraphics[width=\textwidth]{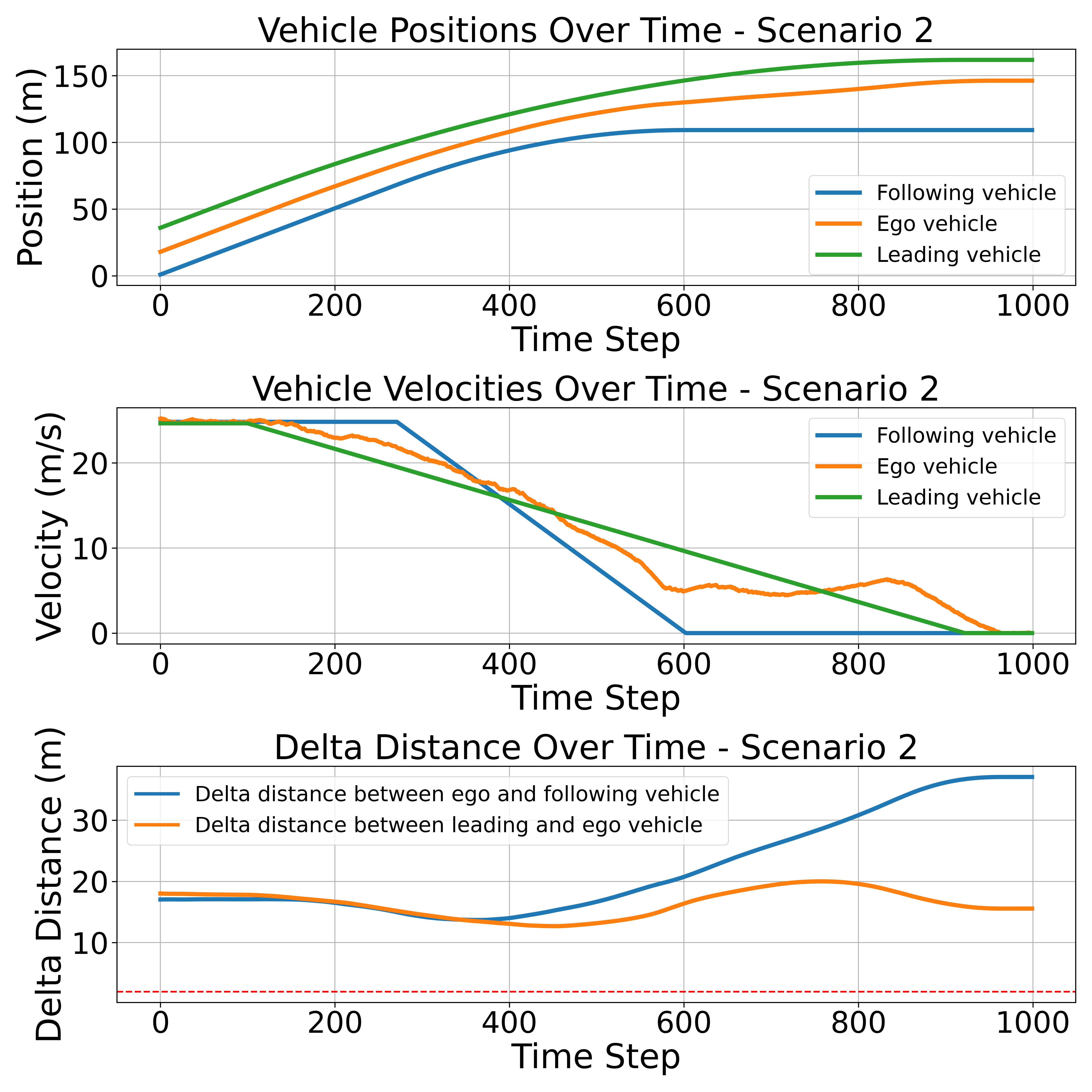}
        \caption{Cut-in scenario 2}
    \end{subfigure}
    \hspace{0mm}
    \begin{subfigure}[t]{0.32\textwidth}
        \centering
        \includegraphics[width=\textwidth]{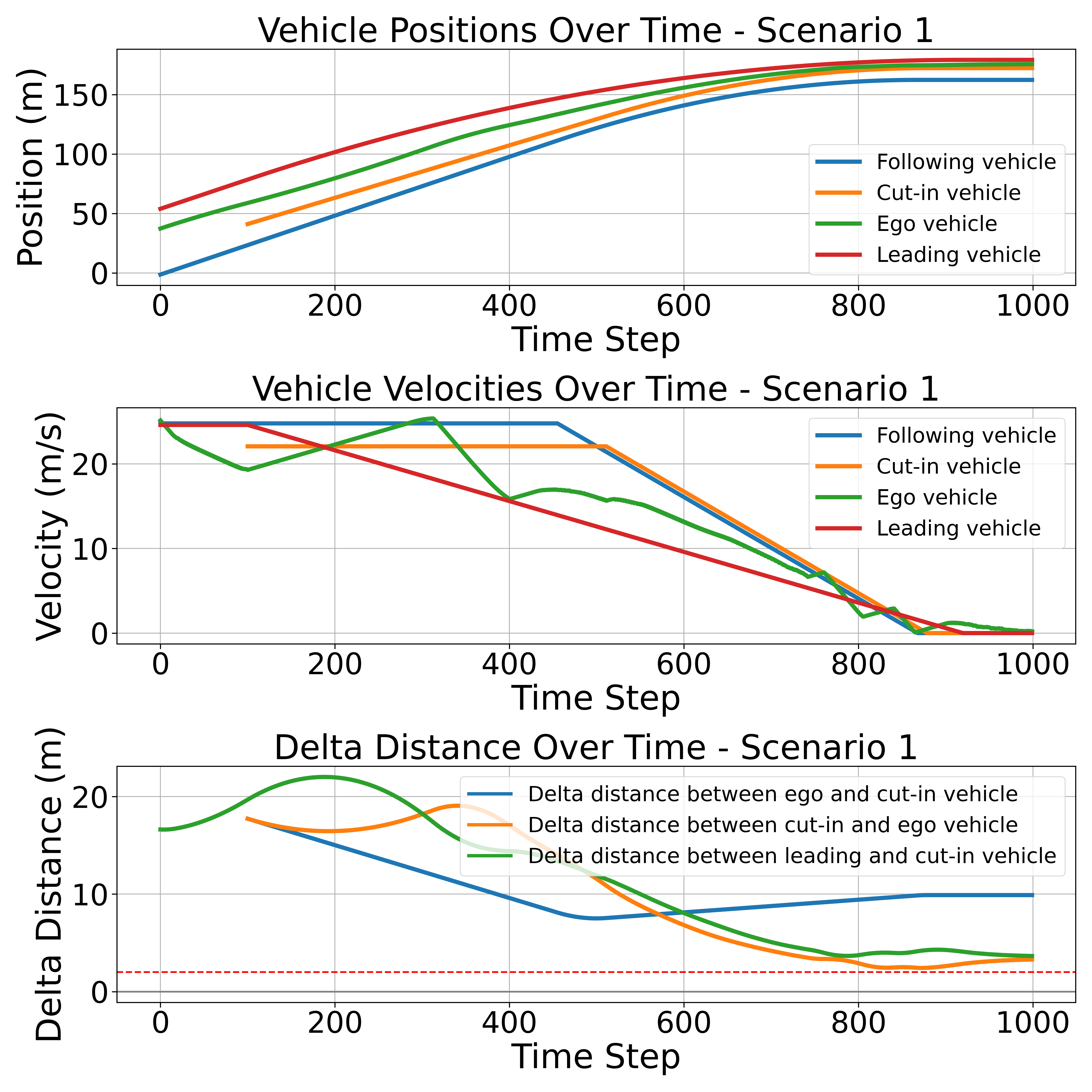}
        \caption{Highway emergency brake scenario 1}
    \end{subfigure}
    \quad
    \begin{subfigure}[t]{0.32\textwidth}
        \centering
        \includegraphics[width=\textwidth]{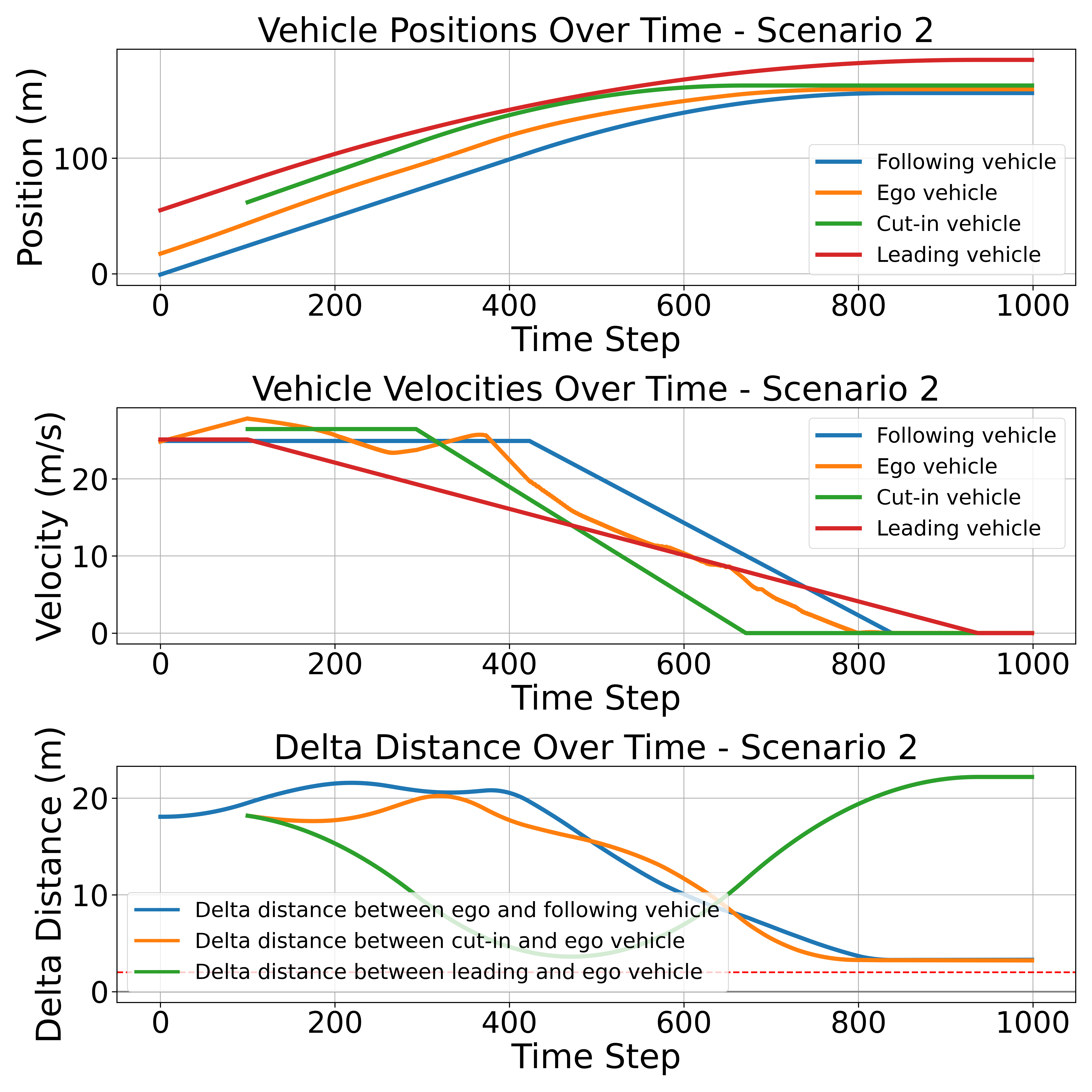}
        \caption{Highway emergency brake scenario 2}
    \end{subfigure}
    \hspace{0mm}
    \begin{subfigure}[t]{0.32\textwidth}
        \centering
        \includegraphics[width=\textwidth]{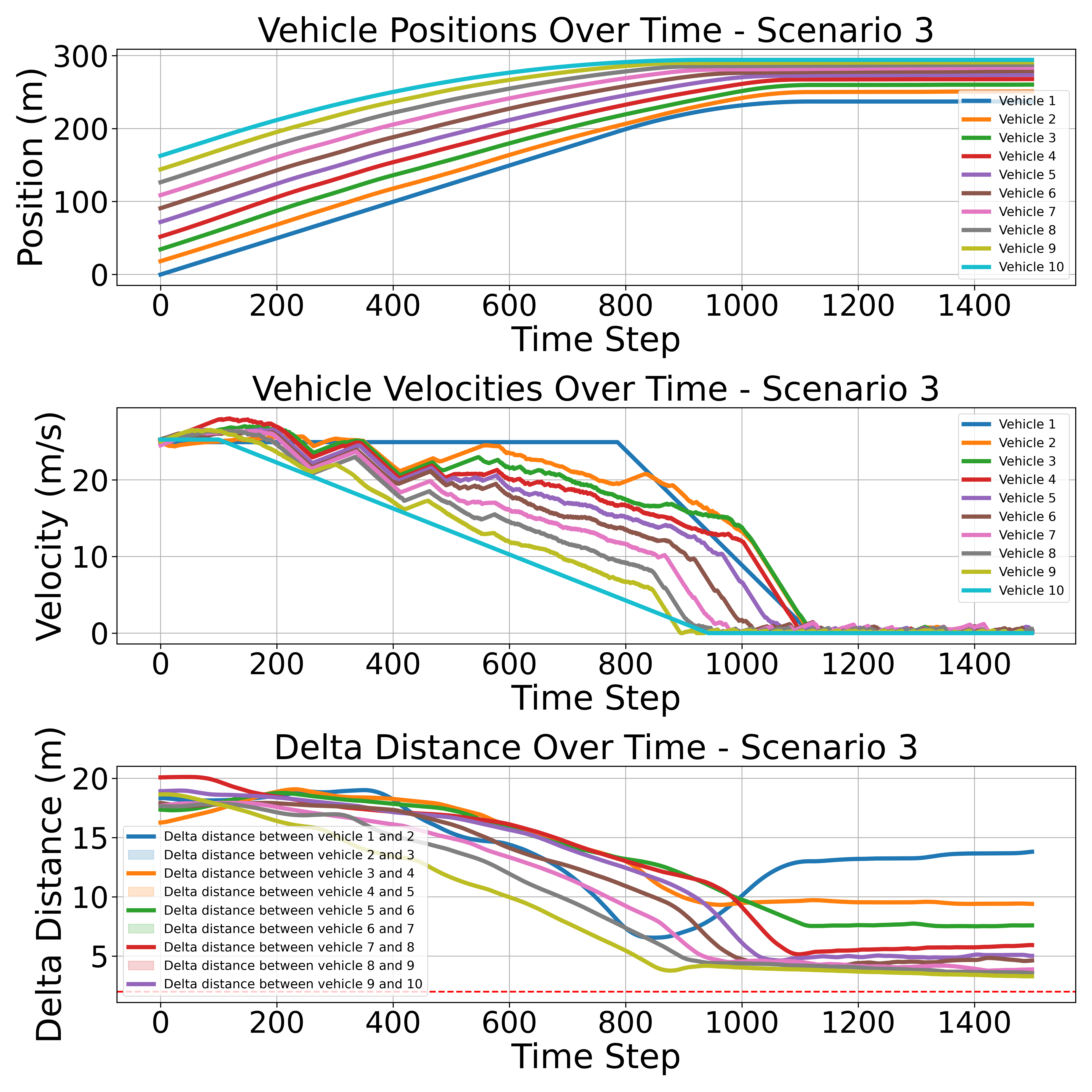}
        \caption{Multi-vehicle emergency brake scenario}
    \end{subfigure}
    
    \caption{Proposed DRL algorithm implementation in proposed edge case scenarios}
    \label{DRLscenario34}
\end{figure}

    
    
Proposed distance-related RL algorithm implementation in scenarios 1 and 2 are shown in Fig.~\ref{DRLscenario34}. In the cut-in scenarios, the cut-in vehicle's velocity and position are the median velocity and median position of the leading and following vehicles. In scenario 1, the red leading vehicle starts to brake at $100^{th}$ time step and the yellow vehicle (truck) starts cut-in in 100 time step. Before the cut-in vehicle (truck) is considered as the following vehicle of the RL vehicle, the green RL vehicle is closer to the red leading vehicle compared to the blue following vehicle. Thus, the green vehicle will try to find the optimal velocity and gap distance between both leading and following vehicles. After the cut-in vehicle is considered as the following vehicle, the green RL vehicle will immediately accelerate even when the leading vehicle is deceleration. At this time step, the green RL vehicle will consider the following vehicle more dangerous because of the combination of the gap distance, velocity, and acceleration. After a few more time steps, the green RL vehicle will start to decelerate more quickly, still choose the optimal action, and avoid the final collision with both the leading and following vehicles.

In scenario 2, the red leading vehicle and the yellow vehicle (sedan) start their action with the same configuration as scenario 1. Before the cut-in vehicle (sedan) was considered the leading vehicle of the RL vehicle, the green RL vehicle was closer to the blue following vehicle compared to the red leading vehicle. Thus, the green vehicle will act the same as scenario 1 to find the optimal velocity and gap distance between both leading and following vehicles. After the cut-in vehicle is considered as the leading vehicle, the green RL vehicle will immediately decelerate even if the following vehicle maintains its velocity. At this time step, the green RL vehicle will consider the cut-in vehicle more dangerous because of the combination of the gap distance, velocity, and acceleration. After a few more time steps, the green RL vehicle will reduce the deceleration, still choose the optimal action, and avoid the final collision with both the leading and following vehicles.

The pattern feature of the well-trained RL algorithm can be significantly found in scenarios 1 and 2. The RL vehicle will not act as the ADAS algorithm, which has an obvious break threshold, and actions that go against the human driver's common sense, such as accelerating when the leading vehicle starts to significantly decelerate, will even be chosen by the RL vehicle.

\subsubsection{Highway emergency brake scenario}

Proposed distance-related RL algorithm implementation in highway emergency brake scenarios are shown in Fig.~\ref{DRLscenario34}.In the highway emergency brake scenarios, the only difference between scenarios 3 and 4 is the deceleration. Compared to the ADAS driving algorithm implementation which causes the collision in scenario 3 and avoids the collision in scenario 4, the proposed RL algorithm will avoid the collision in both scenarios 3 and 4. Because we designed scenario 3 precisely, there will be a one-vehicle length gap between the green leading vehicle (sedan) and the blue following vehicle (truck) always, the proposed RL algorithm will successfully find the optimal deceleration in each time step and finally stop in the gap without having any collision in the procedure.

\subsubsection{Highway multiple RL vehicles following scenario}

The RL vehicle is also implemented in multi-agent scenarios. The result is shown in Fig.~\ref{DRLscenario34}. The RL vehicle is trained as a single agent in a random vehicle following scenario with a 2m safety threshold. At the $100^{th}$ time step, the front leading vehicle (sedan) will decelerate suddenly, and the last following vehicle (truck) will take the AEB decelerate action with a lower maximum deceleration compared to the leading vehicle. This means that the fully stopped gap distance between the front leading vehicle and the last following vehicle will be shorter than the highway vehicle following the gap distance. To verify the multi-vehicle's optimal action, the fully stopped gap distance is precisely defined to be as short as possible while still allowing enough space for the multiple RL vehicles to successfully stop in the gap between the leading and following vehicles without collision. The implementation of the multiple RL vehicles shows that the fully stopped gap distances between each pair of close vehicles are longer than the collision threshold of 2m. It can also be concluded that the RL vehicle will not just respond to the fixed vehicle deceleration and acceleration features but will successfully avoid collisions by adjusting its actions based on the current state of the leading and following vehicles, including their position, velocity, and acceleration.

\subsection{Trajectory calibration model result}
\begin{figure}[htb]
        \centerline{\includegraphics[width=0.35\textwidth]{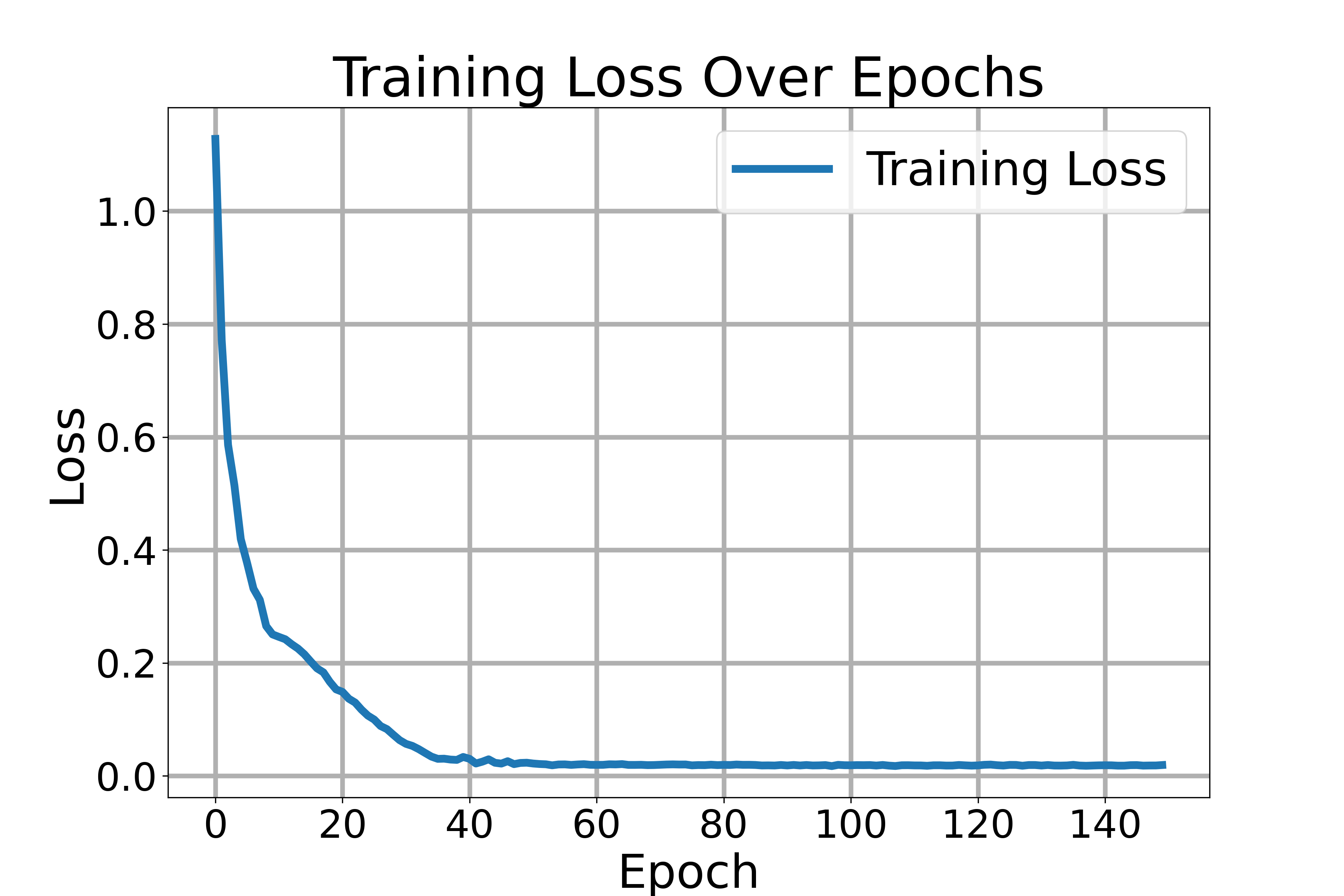}}
        \caption{Trajectory calibration model training loss function}
        \label{Trajectorylossfunction}
    \end{figure}

As designed in the preprocessing model experiment part, 10 rear camera video datasets were acquired. Each dataset is approximately 30 minutes in length, and around 10 clips (each 10 seconds long) that include the process of the rear vehicle approaching or leaving were selected. For one 10-second clip, 100 useful video frames and the related vehicle position in the CARLA map were acquired by the CARLA UE4 engine's physic model. The YOLOv8 (yolov8n.pt) model was used to recognize the vehicle's bounding box and centering with ID and ground truth position. The dashed single white lane's gap distance in CARLA town06 is $7.30m$. After the camera perspective transformation model was established, the bounding box centering dataset was transferred from the rear camera to the top view camera and linked with each frame's ground truth dataset. After acquiring the transformed bounding box centering, the bounding box centering (x, y) was considered as input data, and the ground truth vehicle delta distance $(\Delta x, \Delta y)$ was considered as output data, with the test dataset ratio being 0.2. The loss function is shown in Fig.~\ref{Trajectorylossfunction}.

\begin{figure}[htb]
    \centering
    \begin{subfigure}[b]{0.3\textwidth}
        \centering
        \includegraphics[width=\linewidth]{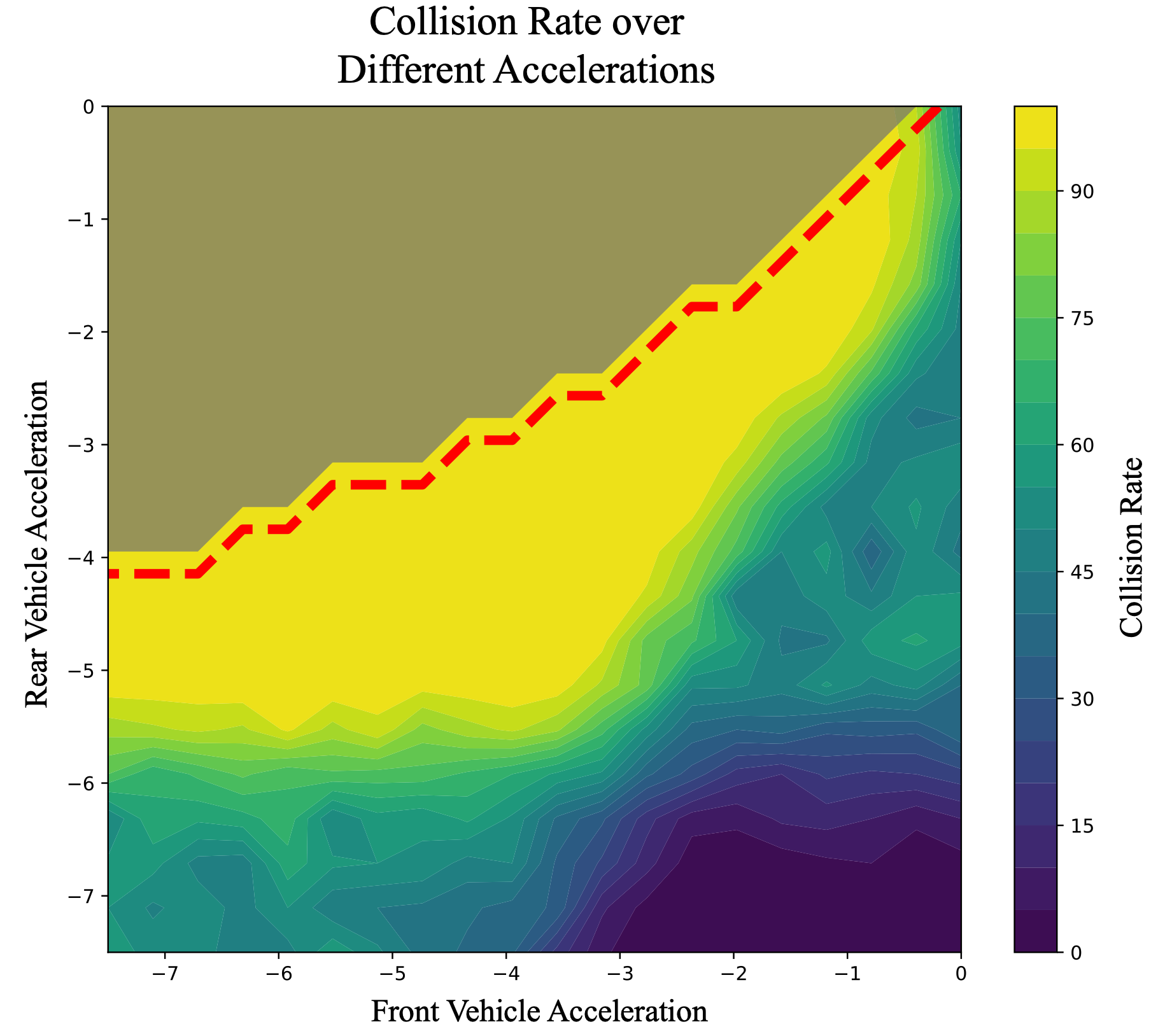}
        \caption{Baseline ADAS driving algorithm validation}
        \label{fig:heatmap1}
    \end{subfigure}
    \begin{subfigure}[b]{0.3\textwidth}
        \centering
        \includegraphics[width=\linewidth]{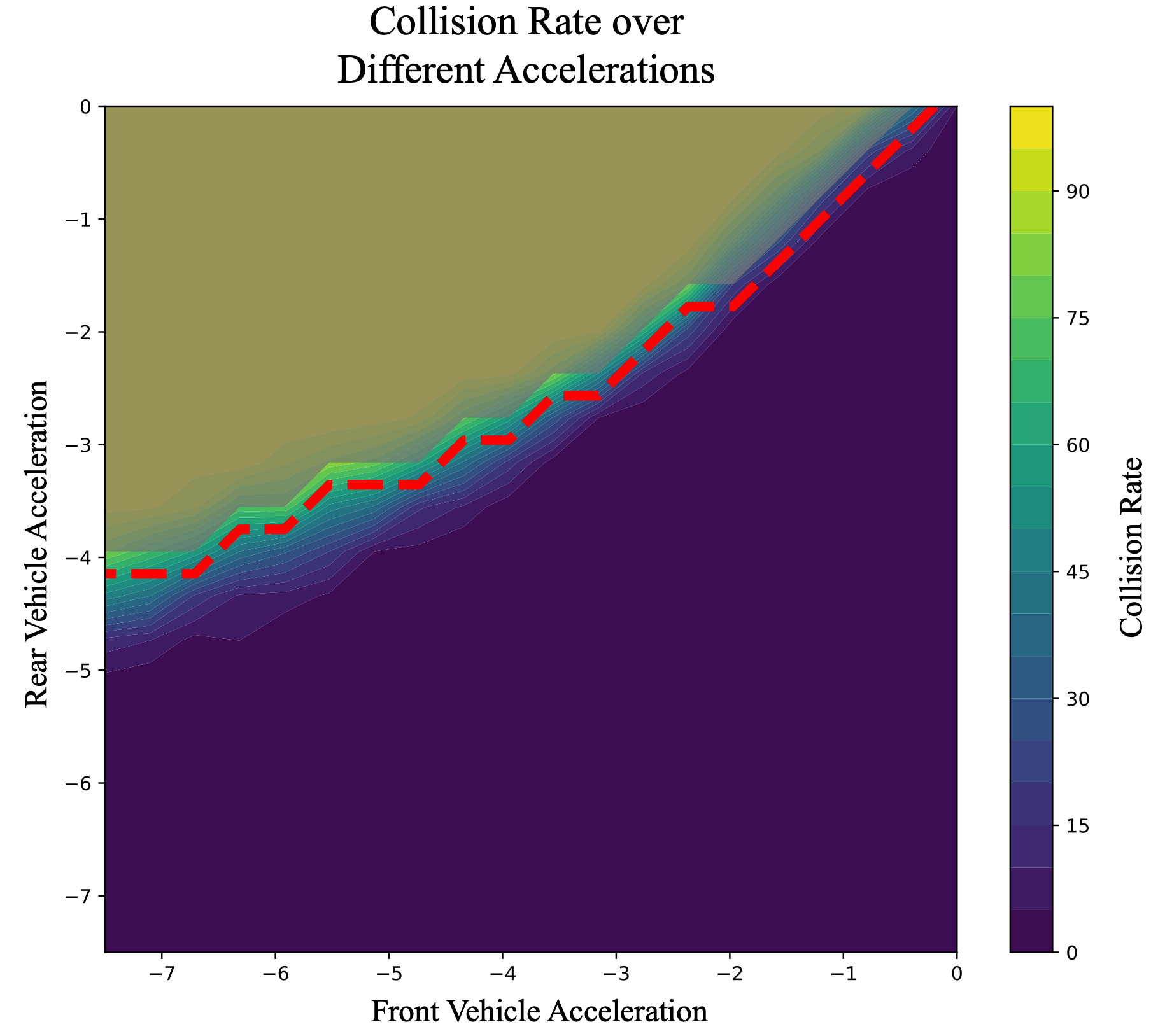}
        \caption{Well-trained proposed RL algorithm validation}
        \label{fig:heatmap2}
    \end{subfigure}
    \begin{subfigure}[b]{0.3\textwidth}
        \centering
        \includegraphics[width=\linewidth]{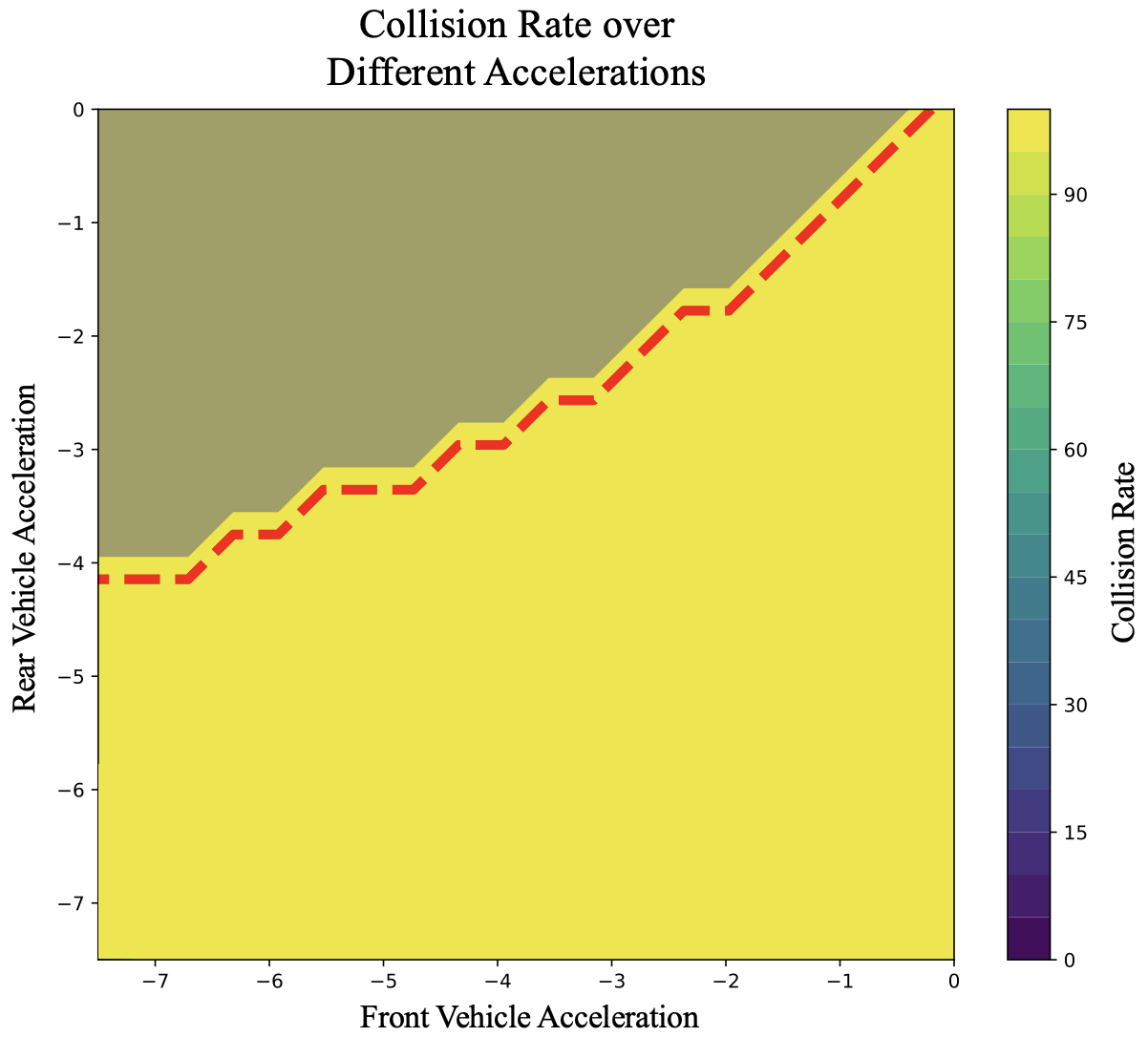}
        \caption{Un-converged proposed RL algorithm validation}
        \label{fig:heatmap3}
    \end{subfigure}
    
    \caption{Algorithm evaluation in different vehicle dynamic patterns.}
    \label{fig:safety_performance}
\end{figure}

\subsection{Algorithm evaluation in different vehicle dynamic patterns}
\begin{figure*}[htbp]
    \centering
    \begin{subfigure}[b]{0.64\textwidth}
        \centering
        \includegraphics[width=\linewidth]{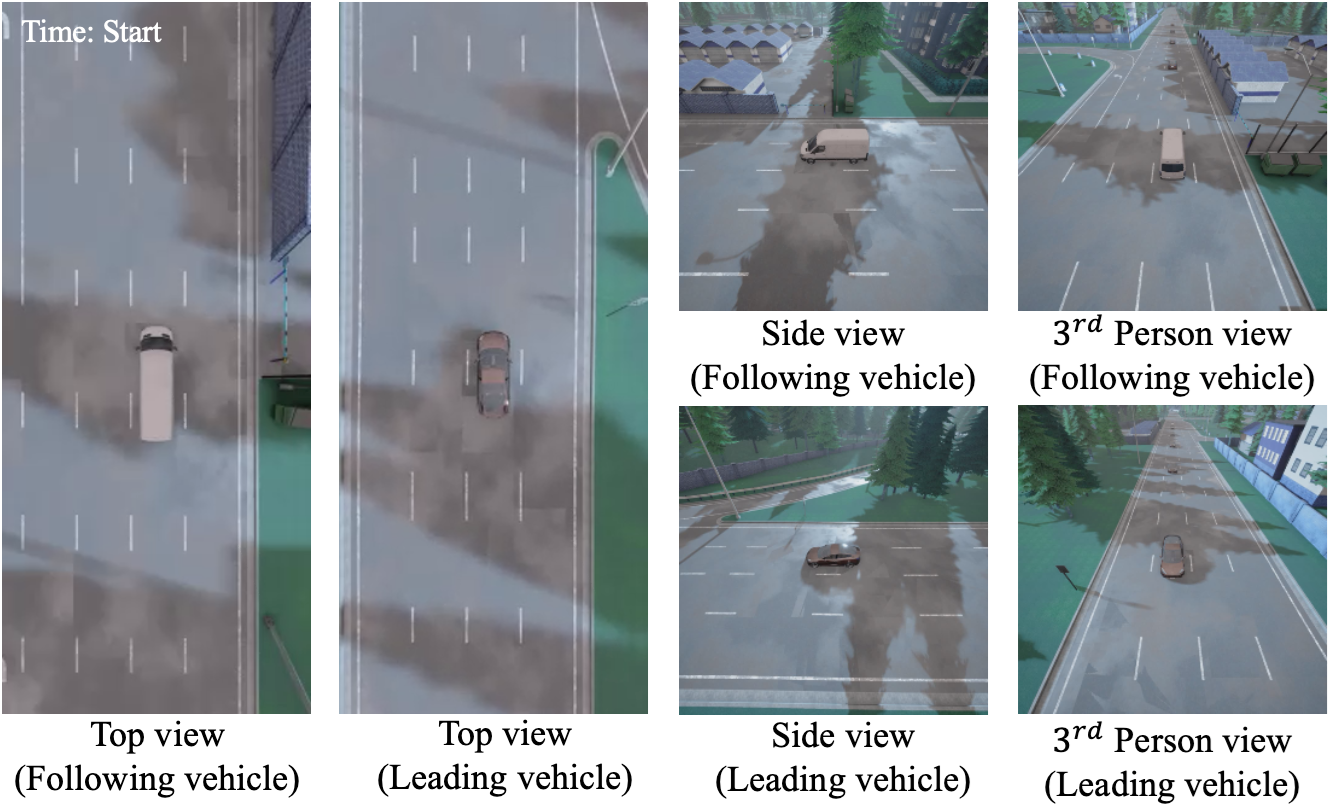}
        \caption{Vehicle positions at the beginning of the scenario}
        \label{fig:carla_view_1}
    \end{subfigure}
    \quad
    \begin{subfigure}[b]{0.64\textwidth}
        \centering
        \includegraphics[width=\linewidth]{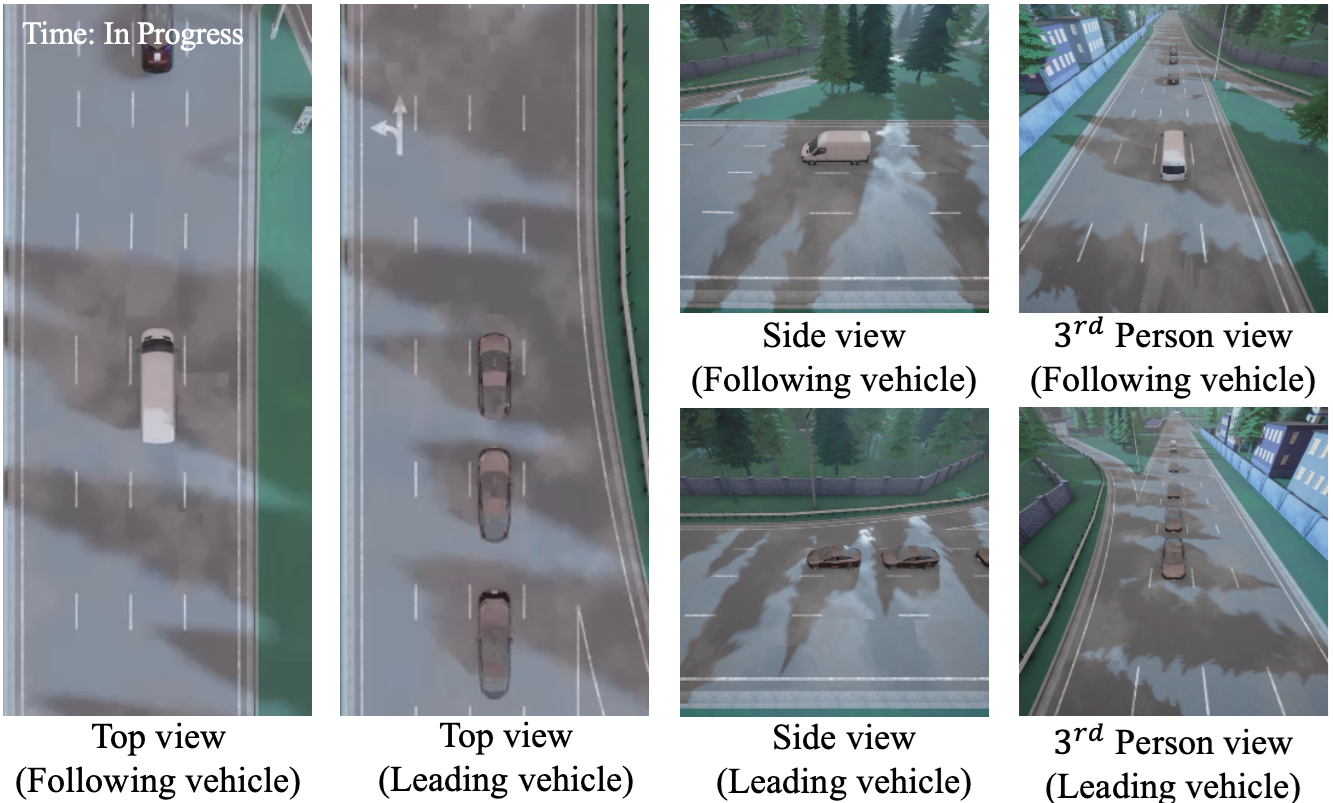}
        \caption{Vehicle positions as the scenario progresses}
        \label{fig:carla_view_2}
    \end{subfigure}
    \quad
    \begin{subfigure}[b]{0.8\textwidth}
        \centering
        \includegraphics[width=\linewidth]{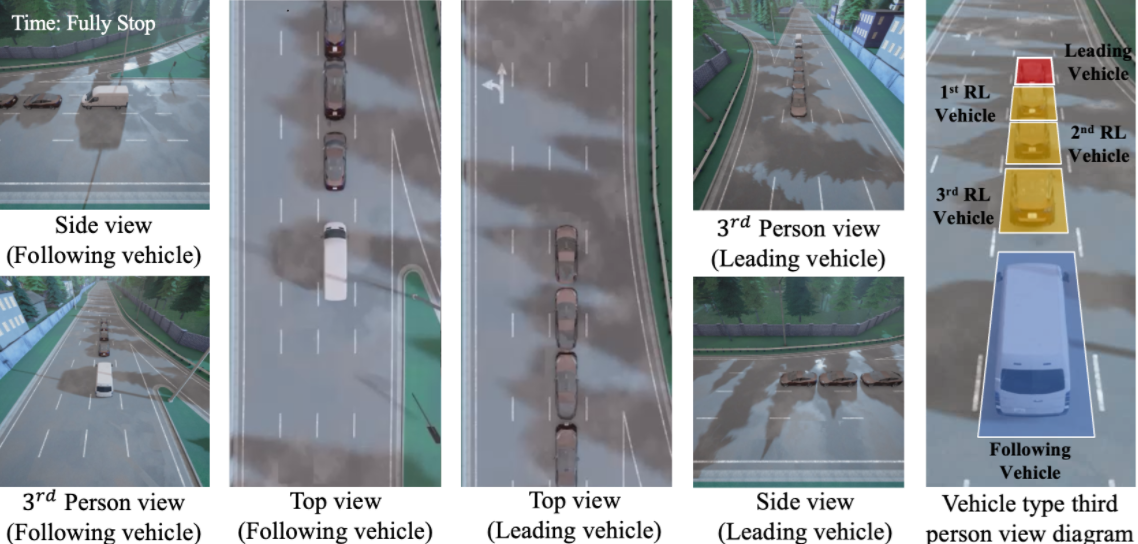}
        \caption{Vehicle positions at the end of the scenario}
        \label{fig:carla_view_3}
    \end{subfigure}
    
    \caption{RL algorithm implementation and verification in CARLA 0.9.13}
    \label{fig:carla_implementation}
\end{figure*}
The well-trained algorithm's safety performance will usually evaluated by the comparison with the baseline algorithm and no ego-vehicle involved scenario. Here we proposed a scenario with only three vehicles involved, the middle vehicle is the RL vehicle, and the leading/following vehicle has the traditional ADAS which combines AEB and ACC algorithms. Both leading and following vehicles' AEB deceleration range is $np.linspace[-7.5m/s^2, 0m/s^2,20]$.Each deceleration combination will be simulated 100 times. The total number of simulation cases is $20*20*100$. The collision rate of the proposed well-trained RL algorithm is shown in Fig.~\ref{fig:safety_performance}. The value represents the percentage of each deceleration combination by calculating the~\ref{eq:probcollision} accumulated collision cases. Also, the nongrey area which is divided by red dash line segmentation in the heatmap represents the theoretically potential no collision cases in an ideal vehicle emergency brake situation (enough space for ego vehicle to avoid collision). In Fig.~\ref{fig:safety_performance}, Subfig.~\ref{fig:heatmap1} represents the baseline ADAS driving algorithm's performance of different vehicle dynamic patterns and the probability of successful avoidance is $36.77\%$. Subfig.~\ref{fig:heatmap2} represents the proposed well-trained RL longitudinal control algorithm's performance of different vehicle dynamic patterns and the probability of successful avoidance is $99\%$. Subfig.~\ref{fig:heatmap3} represents the un-converged RL algorithm's result and it will lead to a crash in any circumstances without successful training.


\subsection{CARLA implementation}

To verify the combination of the trajectory calibration model in the CARLA simulator, a straight-line driving scenario in Town06 was established, and the camera was attached to the leading and following vehicle at the same height and pitch angle. The video frame is output by the camera in each CARLA simulator time step. Synchrony mode is used in the simulator due to a slight delay of a few frames in the camera data. To match the well-trained model with a $2m$ collision threshold, a small vehicle, "Tesla Model 3," is chosen for the platoon, except for the last one with lower deceleration, "Ford Transit". After the combination of the RL model and the trajectory calibration model was implemented, the vehicle platoon was tested with different random initial positions, proving that all RL vehicles could stop successfully without collisions. The implementation screenshot is shown in Fig.~\ref{fig:carla_implementation}.

\section{Conclusions}

In this study, a novel DRL-based reward function is proposed to train vehicle agents to explore optimal actions in various edge-case scenarios. The algorithm considers not only the leading vehicle as part of the environment but also includes the following vehicle. Additionally, a simulator-involved sensor preprocessing model using CARLA is introduced to calibrate the trajectories of both leading and following vehicles, aiming to reduce potential variance and error influences on the RL algorithm.

Implementation in a virtual environment and co-simulator verification shows that the proposed algorithm is safer compared to current ADAS driving algorithms. By utilizing the DDPG model, a significant gap in existing ACC and AEB systems is addressed by considering the behavior of both leading and following vehicles. In simulated high-risk scenarios involving emergency braking in dense traffic, where traditional systems often fail, potential pile-up collisions, including those involving heavy vehicles, are effectively prevented by using the algorithm to raise the no-collision rate from $36.77\%$ to $99\%$. The study contributes to the ongoing evolution of ADAS technology by incorporating artificial intelligence and shifting the paradigm from reactive systems to proactive safety mechanisms. Future work will focus on enhancing the RL algorithm by integrating vehicle and sensor dynamics and physical features, with the goal of applying it to real-life vehicles.

\section{CRediT}
\textbf{Dianwei Chen}: Conceptualization, Methodology, Writing - original draft. \textbf{Yaobang Gong}: Conceptualization, Writing - review \& editing, Experiment design. \textbf{Xianfeng Terry Yang}: Conceptualization, Methodology and Supervision.

\section{Acknowledgment}
This research is supported by the award ”Collaborative Research: OAC Core: Stochastic Simulation Platform for Assessing Safety Performance of Autonomous Vehicles in Winter Seasons (\# 2234292)” which is funded by the National Science Foundation.

\printcredits

\bibliographystyle{cas-model2-names}

\bibliography{cas-sc-template}

\begin{thebibliography}{42}
\expandafter\ifx\csname natexlab\endcsname\relax\def\natexlab#1{#1}\fi
\providecommand{\url}[1]{\texttt{#1}}
\providecommand{\href}[2]{#2}
\providecommand{\path}[1]{#1}
\providecommand{\DOIprefix}{doi:}
\providecommand{\ArXivprefix}{arXiv:}
\providecommand{\URLprefix}{URL: }
\providecommand{\Pubmedprefix}{pmid:}
\providecommand{\doi}[1]{\href{http://dx.doi.org/#1}{\path{#1}}}
\providecommand{\Pubmed}[1]{\href{pmid:#1}{\path{#1}}}
\providecommand{\bibinfo}[2]{#2}
\ifx\xfnm\relax \def\xfnm[#1]{\unskip,\space#1}\fi
\bibitem[{Abbas et~al.(2019)Abbas, O’Kelly, Rodionova and Mangharam}]{abbas2019safe}
\bibinfo{author}{Abbas, H.}, \bibinfo{author}{O’Kelly, M.}, \bibinfo{author}{Rodionova, A.}, \bibinfo{author}{Mangharam, R.}, \bibinfo{year}{2019}.
\newblock \bibinfo{title}{Safe at any speed: A simulation-based test harness for autonomous vehicles}, in: \bibinfo{booktitle}{Cyber Physical Systems. Design, Modeling, and Evaluation: 7th International Workshop, CyPhy 2017, Seoul, South Korea, October 15-20, 2017, Revised Selected Papers 7}, \bibinfo{organization}{Springer}. pp. \bibinfo{pages}{94--106}.
\bibitem[{Burton et~al.(1997)Burton, Delaney, Newstead, Logan and Fildes}]{burton1997evaluation}
\bibinfo{author}{Burton, D.}, \bibinfo{author}{Delaney, A.}, \bibinfo{author}{Newstead, S.}, \bibinfo{author}{Logan, D.}, \bibinfo{author}{Fildes, B.}, \bibinfo{year}{1997}.
\newblock \bibinfo{title}{Evaluation of anti-lock braking systems effectiveness}.
\newblock \bibinfo{journal}{Accident Analysis and Prevention} \bibinfo{volume}{29}, \bibinfo{pages}{745--757}.
\bibitem[{Cafiso and Di~Graziano(2012)}]{cafiso2012evaluation}
\bibinfo{author}{Cafiso, S.}, \bibinfo{author}{Di~Graziano, A.}, \bibinfo{year}{2012}.
\newblock \bibinfo{title}{Evaluation of the effectiveness of adas in reducing multi-vehicle collisions}.
\newblock \bibinfo{journal}{International journal of heavy vehicle systems} \bibinfo{volume}{19}, \bibinfo{pages}{188--206}.
\bibitem[{Chae et~al.(2017)Chae, Kang, Kim, Kim, Chung and Choi}]{chae2017autonomous}
\bibinfo{author}{Chae, H.}, \bibinfo{author}{Kang, C.M.}, \bibinfo{author}{Kim, B.}, \bibinfo{author}{Kim, J.}, \bibinfo{author}{Chung, C.C.}, \bibinfo{author}{Choi, J.W.}, \bibinfo{year}{2017}.
\newblock \bibinfo{title}{Autonomous braking system via deep reinforcement learning}, in: \bibinfo{booktitle}{2017 IEEE 20th International conference on intelligent transportation systems (ITSC)}, \bibinfo{organization}{IEEE}. pp. \bibinfo{pages}{1--6}.
\bibitem[{Chen et~al.(2024)Chen, Gong and Yang}]{chen2024deep}
\bibinfo{author}{Chen, D.}, \bibinfo{author}{Gong, Y.}, \bibinfo{author}{Yang, X.}, \bibinfo{year}{2024}.
\newblock \bibinfo{title}{Deep reinforcement learning for advanced longitudinal control and collision avoidance in high-risk driving scenarios}.
\newblock \bibinfo{journal}{arXiv preprint arXiv:2404.19087} .
\bibitem[{Chen et~al.(2023)Chen, Yurtsever, Redmill and {\"O}zg{\"u}ner}]{chen2023using}
\bibinfo{author}{Chen, D.}, \bibinfo{author}{Yurtsever, E.}, \bibinfo{author}{Redmill, K.A.}, \bibinfo{author}{{\"O}zg{\"u}ner, {\"U}.}, \bibinfo{year}{2023}.
\newblock \bibinfo{title}{Using collision momentum in deep reinforcement learning based adversarial pedestrian modeling}, in: \bibinfo{booktitle}{2023 IEEE Intelligent Vehicles Symposium (IV)}, \bibinfo{organization}{IEEE}. pp. \bibinfo{pages}{1--6}.
\bibitem[{Chen et~al.(2025)Chen, Zhang, Liu and Yang}]{chen2025insight}
\bibinfo{author}{Chen, D.}, \bibinfo{author}{Zhang, Z.}, \bibinfo{author}{Liu, Y.}, \bibinfo{author}{Yang, X.T.}, \bibinfo{year}{2025}.
\newblock \bibinfo{title}{Insight: Enhancing autonomous driving safety through vision-language models on context-aware hazard detection and edge case evaluation}.
\newblock \bibinfo{journal}{arXiv e-prints} , \bibinfo{pages}{arXiv--2502}.
\bibitem[{De~Wit and Tsiotras(1999)}]{de1999dynamic}
\bibinfo{author}{De~Wit, C.C.}, \bibinfo{author}{Tsiotras, P.}, \bibinfo{year}{1999}.
\newblock \bibinfo{title}{Dynamic tire friction models for vehicle traction control}, in: \bibinfo{booktitle}{Proceedings of the 38th IEEE conference on decision and control (Cat. no. 99CH36304)}, \bibinfo{organization}{IEEE}. pp. \bibinfo{pages}{3746--3751}.
\bibitem[{Desjardins and Chaib-Draa(2011)}]{desjardins2011cooperative}
\bibinfo{author}{Desjardins, C.}, \bibinfo{author}{Chaib-Draa, B.}, \bibinfo{year}{2011}.
\newblock \bibinfo{title}{Cooperative adaptive cruise control: A reinforcement learning approach}.
\newblock \bibinfo{journal}{IEEE Transactions on intelligent transportation systems} \bibinfo{volume}{12}, \bibinfo{pages}{1248--1260}.
\bibitem[{Donnell et~al.(2009)Donnell, Hines, Mahoney, Porter, McGee et~al.}]{donnell2009speed}
\bibinfo{author}{Donnell, E.T.}, \bibinfo{author}{Hines, S.C.}, \bibinfo{author}{Mahoney, K.M.}, \bibinfo{author}{Porter, R.J.}, \bibinfo{author}{McGee, H.}, et~al., \bibinfo{year}{2009}.
\newblock \bibinfo{title}{Speed concepts: Informational guide}.
\newblock \bibinfo{type}{Technical Report}. United States. Federal Highway Administration. Office of Safety.
\bibitem[{Dosovitskiy et~al.(2017)Dosovitskiy, Ros, Codevilla, Lopez and Koltun}]{dosovitskiy2017carla}
\bibinfo{author}{Dosovitskiy, A.}, \bibinfo{author}{Ros, G.}, \bibinfo{author}{Codevilla, F.}, \bibinfo{author}{Lopez, A.}, \bibinfo{author}{Koltun, V.}, \bibinfo{year}{2017}.
\newblock \bibinfo{title}{Carla: An open urban driving simulator}, in: \bibinfo{booktitle}{Conference on robot learning}, \bibinfo{organization}{PMLR}. pp. \bibinfo{pages}{1--16}.
\bibitem[{Feng et~al.(2023)Feng, Sun, Yan, Zhu, Zou, Shen and Liu}]{feng2023dense}
\bibinfo{author}{Feng, S.}, \bibinfo{author}{Sun, H.}, \bibinfo{author}{Yan, X.}, \bibinfo{author}{Zhu, H.}, \bibinfo{author}{Zou, Z.}, \bibinfo{author}{Shen, S.}, \bibinfo{author}{Liu, H.X.}, \bibinfo{year}{2023}.
\newblock \bibinfo{title}{Dense reinforcement learning for safety validation of autonomous vehicles}.
\newblock \bibinfo{journal}{Nature} \bibinfo{volume}{615}, \bibinfo{pages}{620--627}.
\bibitem[{Ferguson(2007)}]{ferguson2007effectiveness}
\bibinfo{author}{Ferguson, S.A.}, \bibinfo{year}{2007}.
\newblock \bibinfo{title}{The effectiveness of electronic stability control in reducing real-world crashes: a literature review}.
\newblock \bibinfo{journal}{Traffic injury prevention} \bibinfo{volume}{8}, \bibinfo{pages}{329--338}.
\bibitem[{Fildes et~al.(2015)Fildes, Keall, Bos, Lie, Page, Pastor, Pennisi, Rizzi, Thomas and Tingvall}]{fildes2015effectiveness}
\bibinfo{author}{Fildes, B.}, \bibinfo{author}{Keall, M.}, \bibinfo{author}{Bos, N.}, \bibinfo{author}{Lie, A.}, \bibinfo{author}{Page, Y.}, \bibinfo{author}{Pastor, C.}, \bibinfo{author}{Pennisi, L.}, \bibinfo{author}{Rizzi, M.}, \bibinfo{author}{Thomas, P.}, \bibinfo{author}{Tingvall, C.}, \bibinfo{year}{2015}.
\newblock \bibinfo{title}{Effectiveness of low speed autonomous emergency braking in real-world rear-end crashes}.
\newblock \bibinfo{journal}{Accident Analysis \& Prevention} \bibinfo{volume}{81}, \bibinfo{pages}{24--29}.
\bibitem[{Fu et~al.(2020)Fu, Li, Yu, Luan and Zhang}]{fu2020decision}
\bibinfo{author}{Fu, Y.}, \bibinfo{author}{Li, C.}, \bibinfo{author}{Yu, F.R.}, \bibinfo{author}{Luan, T.H.}, \bibinfo{author}{Zhang, Y.}, \bibinfo{year}{2020}.
\newblock \bibinfo{title}{A decision-making strategy for vehicle autonomous braking in emergency via deep reinforcement learning}.
\newblock \bibinfo{journal}{IEEE transactions on vehicular technology} \bibinfo{volume}{69}, \bibinfo{pages}{5876--5888}.
\bibitem[{Galvani(2019)}]{galvani2019history}
\bibinfo{author}{Galvani, M.}, \bibinfo{year}{2019}.
\newblock \bibinfo{title}{History and future of driver assistance}.
\newblock \bibinfo{journal}{IEEE Instrumentation \& Measurement Magazine} \bibinfo{volume}{22}, \bibinfo{pages}{11--16}.
\bibitem[{Guo et~al.(2019)Guo, Kurup and Shah}]{guo2019safe}
\bibinfo{author}{Guo, J.}, \bibinfo{author}{Kurup, U.}, \bibinfo{author}{Shah, M.}, \bibinfo{year}{2019}.
\newblock \bibinfo{title}{Is it safe to drive? an overview of factors, metrics, and datasets for driveability assessment in autonomous driving}.
\newblock \bibinfo{journal}{IEEE Transactions on Intelligent Transportation Systems} \bibinfo{volume}{21}, \bibinfo{pages}{3135--3151}.
\bibitem[{Kang and Kum(2020)}]{kang2020camera}
\bibinfo{author}{Kang, D.}, \bibinfo{author}{Kum, D.}, \bibinfo{year}{2020}.
\newblock \bibinfo{title}{Camera and radar sensor fusion for robust vehicle localization via vehicle part localization}.
\newblock \bibinfo{journal}{IEEE Access} \bibinfo{volume}{8}, \bibinfo{pages}{75223--75236}.
\bibitem[{Kim et~al.(2017)Kim, Bong, Park and Park}]{kim2017prediction}
\bibinfo{author}{Kim, I.H.}, \bibinfo{author}{Bong, J.H.}, \bibinfo{author}{Park, J.}, \bibinfo{author}{Park, S.}, \bibinfo{year}{2017}.
\newblock \bibinfo{title}{Prediction of driver’s intention of lane change by augmenting sensor information using machine learning techniques}.
\newblock \bibinfo{journal}{Sensors} \bibinfo{volume}{17}, \bibinfo{pages}{1350}.
\bibitem[{Koopman and Wagner(2016)}]{koopman2016challenges}
\bibinfo{author}{Koopman, P.}, \bibinfo{author}{Wagner, M.}, \bibinfo{year}{2016}.
\newblock \bibinfo{title}{Challenges in autonomous vehicle testing and validation}.
\newblock \bibinfo{journal}{SAE International Journal of Transportation Safety} \bibinfo{volume}{4}, \bibinfo{pages}{15--24}.
\bibitem[{Kozak et~al.(2006)Kozak, Pohl, Birk, Greenberg, Artz, Blommer, Cathey and Curry}]{kozak2006evaluation}
\bibinfo{author}{Kozak, K.}, \bibinfo{author}{Pohl, J.}, \bibinfo{author}{Birk, W.}, \bibinfo{author}{Greenberg, J.}, \bibinfo{author}{Artz, B.}, \bibinfo{author}{Blommer, M.}, \bibinfo{author}{Cathey, L.}, \bibinfo{author}{Curry, R.}, \bibinfo{year}{2006}.
\newblock \bibinfo{title}{Evaluation of lane departure warnings for drowsy drivers}, in: \bibinfo{booktitle}{Proceedings of the human factors and ergonomics society annual meeting}, \bibinfo{organization}{Sage Publications Sage CA: Los Angeles, CA}. pp. \bibinfo{pages}{2400--2404}.
\bibitem[{Kukkala et~al.(2018)Kukkala, Tunnell, Pasricha and Bradley}]{kukkala2018advanced}
\bibinfo{author}{Kukkala, V.K.}, \bibinfo{author}{Tunnell, J.}, \bibinfo{author}{Pasricha, S.}, \bibinfo{author}{Bradley, T.}, \bibinfo{year}{2018}.
\newblock \bibinfo{title}{Advanced driver-assistance systems: A path toward autonomous vehicles}.
\newblock \bibinfo{journal}{IEEE Consumer Electronics Magazine} \bibinfo{volume}{7}, \bibinfo{pages}{18--25}.
\bibitem[{Leiman(2021)}]{leiman2021law}
\bibinfo{author}{Leiman, T.}, \bibinfo{year}{2021}.
\newblock \bibinfo{title}{Law and tech collide: Foreseeability, reasonableness and advanced driver assistance systems}.
\newblock \bibinfo{journal}{Policy and society} \bibinfo{volume}{40}, \bibinfo{pages}{250--271}.
\bibitem[{Li et~al.(2021)Li, Yang, Zhang, Qu, Cao, Cheng and Li}]{li2021risk}
\bibinfo{author}{Li, G.}, \bibinfo{author}{Yang, Y.}, \bibinfo{author}{Zhang, T.}, \bibinfo{author}{Qu, X.}, \bibinfo{author}{Cao, D.}, \bibinfo{author}{Cheng, B.}, \bibinfo{author}{Li, K.}, \bibinfo{year}{2021}.
\newblock \bibinfo{title}{Risk assessment based collision avoidance decision-making for autonomous vehicles in multi-scenarios}.
\newblock \bibinfo{journal}{Transportation research part C: emerging technologies} \bibinfo{volume}{122}, \bibinfo{pages}{102820}.
\bibitem[{Liu et~al.(2017)Liu, Wang and Zou}]{liu2017radar}
\bibinfo{author}{Liu, G.}, \bibinfo{author}{Wang, L.}, \bibinfo{author}{Zou, S.}, \bibinfo{year}{2017}.
\newblock \bibinfo{title}{A radar-based blind spot detection and warning system for driver assistance}, in: \bibinfo{booktitle}{2017 IEEE 2nd Advanced Information Technology, Electronic and Automation Control Conference (IAEAC)}, \bibinfo{organization}{IEEE}. pp. \bibinfo{pages}{2204--2208}.
\bibitem[{Lu et~al.(2019)Lu, Dong and Hu}]{lu2019energy}
\bibinfo{author}{Lu, C.}, \bibinfo{author}{Dong, J.}, \bibinfo{author}{Hu, L.}, \bibinfo{year}{2019}.
\newblock \bibinfo{title}{Energy-efficient adaptive cruise control for electric connected and autonomous vehicles}.
\newblock \bibinfo{journal}{IEEE Intelligent Transportation Systems Magazine} \bibinfo{volume}{11}, \bibinfo{pages}{42--55}.
\bibitem[{Luo et~al.(2024)Luo, Xu, Lai, Chen, Zhang and Yu}]{luo2024survey}
\bibinfo{author}{Luo, F.M.}, \bibinfo{author}{Xu, T.}, \bibinfo{author}{Lai, H.}, \bibinfo{author}{Chen, X.H.}, \bibinfo{author}{Zhang, W.}, \bibinfo{author}{Yu, Y.}, \bibinfo{year}{2024}.
\newblock \bibinfo{title}{A survey on model-based reinforcement learning}.
\newblock \bibinfo{journal}{Science China Information Sciences} \bibinfo{volume}{67}, \bibinfo{pages}{121101}.
\bibitem[{Makridis et~al.(2020)Makridis, Mattas and Ciuffo}]{8884686}
\bibinfo{author}{Makridis, M.}, \bibinfo{author}{Mattas, K.}, \bibinfo{author}{Ciuffo, B.}, \bibinfo{year}{2020}.
\newblock \bibinfo{title}{Response time and time headway of an adaptive cruise control. an empirical characterization and potential impacts on road capacity}.
\newblock \bibinfo{journal}{IEEE Transactions on Intelligent Transportation Systems} \bibinfo{volume}{21}, \bibinfo{pages}{1677--1686}.
\newblock \DOIprefix\doi{10.1109/TITS.2019.2948646}.
\bibitem[{Moujahid et~al.(2018a)Moujahid, ElAraki~Tantaoui, Hina, Soukane, Ortalda, ElKhadimi and Ramdane-Cherif}]{8441758}
\bibinfo{author}{Moujahid, A.}, \bibinfo{author}{ElAraki~Tantaoui, M.}, \bibinfo{author}{Hina, M.D.}, \bibinfo{author}{Soukane, A.}, \bibinfo{author}{Ortalda, A.}, \bibinfo{author}{ElKhadimi, A.}, \bibinfo{author}{Ramdane-Cherif, A.}, \bibinfo{year}{2018}a.
\newblock \bibinfo{title}{Machine learning techniques in adas: A review}, in: \bibinfo{booktitle}{2018 International Conference on Advances in Computing and Communication Engineering (ICACCE)}, pp. \bibinfo{pages}{235--242}.
\newblock \DOIprefix\doi{10.1109/ICACCE.2018.8441758}.
\bibitem[{Moujahid et~al.(2018b)Moujahid, Tantaoui, Hina, Soukane, Ortalda, ElKhadimi and Ramdane-Cherif}]{moujahid2018machine}
\bibinfo{author}{Moujahid, A.}, \bibinfo{author}{Tantaoui, M.E.}, \bibinfo{author}{Hina, M.D.}, \bibinfo{author}{Soukane, A.}, \bibinfo{author}{Ortalda, A.}, \bibinfo{author}{ElKhadimi, A.}, \bibinfo{author}{Ramdane-Cherif, A.}, \bibinfo{year}{2018}b.
\newblock \bibinfo{title}{Machine learning techniques in adas: A review}, in: \bibinfo{booktitle}{2018 International Conference on Advances in Computing and Communication Engineering (ICACCE)}, \bibinfo{organization}{IEEE}. pp. \bibinfo{pages}{235--242}.
\bibitem[{{National Highway Traffic Safety Administration}(2023)}]{FR2023}
\bibinfo{author}{{National Highway Traffic Safety Administration}}, \bibinfo{year}{2023}.
\newblock \bibinfo{title}{Federal motor vehicle safety standards: Automatic emergency braking systems for light vehicles}.
\newblock \bibinfo{howpublished}{Federal Register}.
\newblock \bibinfo{note}{Available from: ProQuest® Congressional; Accessed: 4/23/2024}.
\bibitem[{Nidamanuri et~al.(2021)Nidamanuri, Nibhanupudi, Assfalg and Venkataraman}]{nidamanuri2021progressive}
\bibinfo{author}{Nidamanuri, J.}, \bibinfo{author}{Nibhanupudi, C.}, \bibinfo{author}{Assfalg, R.}, \bibinfo{author}{Venkataraman, H.}, \bibinfo{year}{2021}.
\newblock \bibinfo{title}{A progressive review: Emerging technologies for adas driven solutions}.
\newblock \bibinfo{journal}{IEEE Transactions on Intelligent Vehicles} \bibinfo{volume}{7}, \bibinfo{pages}{326--341}.
\bibitem[{Okuda et~al.(2014)Okuda, Kajiwara and Terashima}]{okuda2014survey}
\bibinfo{author}{Okuda, R.}, \bibinfo{author}{Kajiwara, Y.}, \bibinfo{author}{Terashima, K.}, \bibinfo{year}{2014}.
\newblock \bibinfo{title}{A survey of technical trend of adas and autonomous driving}, in: \bibinfo{booktitle}{Technical Papers of 2014 International Symposium on VLSI Design, Automation and Test}, \bibinfo{organization}{IEEE}. pp. \bibinfo{pages}{1--4}.
\bibitem[{Sallab et~al.(2016)Sallab, Abdou, Perot and Yogamani}]{sallab2016end}
\bibinfo{author}{Sallab, A.E.}, \bibinfo{author}{Abdou, M.}, \bibinfo{author}{Perot, E.}, \bibinfo{author}{Yogamani, S.}, \bibinfo{year}{2016}.
\newblock \bibinfo{title}{End-to-end deep reinforcement learning for lane keeping assist}.
\newblock \bibinfo{journal}{arXiv preprint arXiv:1612.04340} .
\bibitem[{Sotelo and Barriga(2008)}]{sotelo2008blind}
\bibinfo{author}{Sotelo, M.{\'A}.}, \bibinfo{author}{Barriga, J.}, \bibinfo{year}{2008}.
\newblock \bibinfo{title}{Blind spot detection using vision for automotive applications}.
\newblock \bibinfo{journal}{Journal of Zhejiang university-SCIENCE A} \bibinfo{volume}{9}, \bibinfo{pages}{1369--1372}.
\bibitem[{Vahidi and Eskandarian(2003)}]{vahidi2003research}
\bibinfo{author}{Vahidi, A.}, \bibinfo{author}{Eskandarian, A.}, \bibinfo{year}{2003}.
\newblock \bibinfo{title}{Research advances in intelligent collision avoidance and adaptive cruise control}.
\newblock \bibinfo{journal}{IEEE transactions on intelligent transportation systems} \bibinfo{volume}{4}, \bibinfo{pages}{143--153}.
\bibitem[{Wang et~al.(2022)Wang, Wang, Liang, Zhao, Huang, Xu, Dai and Miao}]{wang2022deep}
\bibinfo{author}{Wang, X.}, \bibinfo{author}{Wang, S.}, \bibinfo{author}{Liang, X.}, \bibinfo{author}{Zhao, D.}, \bibinfo{author}{Huang, J.}, \bibinfo{author}{Xu, X.}, \bibinfo{author}{Dai, B.}, \bibinfo{author}{Miao, Q.}, \bibinfo{year}{2022}.
\newblock \bibinfo{title}{Deep reinforcement learning: A survey}.
\newblock \bibinfo{journal}{IEEE Transactions on Neural Networks and Learning Systems} .
\bibitem[{Wang et~al.(2024)Wang, Yan, Wei, Wang, Bo and Xiao}]{wang2024research}
\bibinfo{author}{Wang, Z.}, \bibinfo{author}{Yan, H.}, \bibinfo{author}{Wei, C.}, \bibinfo{author}{Wang, J.}, \bibinfo{author}{Bo, S.}, \bibinfo{author}{Xiao, M.}, \bibinfo{year}{2024}.
\newblock \bibinfo{title}{Research on autonomous driving decision-making strategies based deep reinforcement learning}, in: \bibinfo{booktitle}{Proceedings of the 2024 4th International Conference on Internet of Things and Machine Learning}, pp. \bibinfo{pages}{211--215}.
\bibitem[{Yan et~al.(2024)Yan, Huang and He}]{yan2024comparison}
\bibinfo{author}{Yan, S.}, \bibinfo{author}{Huang, C.}, \bibinfo{author}{He, D.}, \bibinfo{year}{2024}.
\newblock \bibinfo{title}{A comparison of patterns and contributing factors of adas and ads involved crashes}.
\newblock \bibinfo{journal}{Journal of Transportation Safety \& Security} \bibinfo{volume}{16}, \bibinfo{pages}{1061--1088}.
\bibitem[{Yang et~al.(2022)Yang, Yang, Wu, Zhao, Fang, Liao, Wang, Zhang et~al.}]{yang2022systematic}
\bibinfo{author}{Yang, L.}, \bibinfo{author}{Yang, Y.}, \bibinfo{author}{Wu, G.}, \bibinfo{author}{Zhao, X.}, \bibinfo{author}{Fang, S.}, \bibinfo{author}{Liao, X.}, \bibinfo{author}{Wang, R.}, \bibinfo{author}{Zhang, M.}, et~al., \bibinfo{year}{2022}.
\newblock \bibinfo{title}{A systematic review of autonomous emergency braking system: impact factor, technology, and performance evaluation}.
\newblock \bibinfo{journal}{Journal of advanced transportation} \bibinfo{volume}{2022}.
\bibitem[{Yu and Wang(2022)}]{yu2022researches}
\bibinfo{author}{Yu, L.}, \bibinfo{author}{Wang, R.}, \bibinfo{year}{2022}.
\newblock \bibinfo{title}{Researches on adaptive cruise control system: A state of the art review}.
\newblock \bibinfo{journal}{Proceedings of the Institution of Mechanical Engineers, Part D: Journal of Automobile Engineering} \bibinfo{volume}{236}, \bibinfo{pages}{211--240}.
\bibitem[{Zhou et~al.(2020)Zhou, Chen and Zou}]{9106866}
\bibinfo{author}{Zhou, T.}, \bibinfo{author}{Chen, M.}, \bibinfo{author}{Zou, J.}, \bibinfo{year}{2020}.
\newblock \bibinfo{title}{Reinforcement learning based data fusion method for multi-sensors}.
\newblock \bibinfo{journal}{IEEE/CAA Journal of Automatica Sinica} \bibinfo{volume}{7}, \bibinfo{pages}{1489--1497}.
\newblock \DOIprefix\doi{10.1109/JAS.2020.1003180}.

\end{thebibliography}

\end{document}